%% file: main.tex
\documentclass[table]{gtech}
\PassOptionsToPackage{table, usenames, dvipsnames}{xcolor}
\usepackage{xcolor}
\usepackage{amssymb}
\usepackage{multirow}
\usepackage{bigdelim}
\usepackage{longtable}
\usepackage{tabularray}
\usepackage{wrapfig}
\usepackage[most]{tcolorbox}
\usepackage{url}
\usepackage{float}
\usepackage{datatool}
\usepackage{enumitem}
\usepackage{subcaption} %
\usepackage[justification=centering]{caption}

\RequirePackage{tgpagella} %
\RequirePackage{mathpazo}  %
\RequirePackage{inconsolata} %
\usepackage{makecell}

\usepackage{adjustbox}
\usepackage{tablefootnote}
\usepackage{threeparttable}

\usepackage{booktabs} %
\usepackage{array}    %
\newcolumntype{C}[1]{>{\centering\arraybackslash}m{#1}}
\newcommand{\splitcell}[2][c]{\makecell[#1]{#2}}

\usepackage[utf8]{inputenc} %
\usepackage[T1]{fontenc}    %
\usepackage{hyperref}       %
\usepackage{url}            %
\usepackage{booktabs}       %
\usepackage{amsfonts}       %
\usepackage{nicefrac}       %
\usepackage{microtype}      %
\usepackage{xcolor}         %
\usepackage{xspace}
\usepackage{amsthm}   %
\usepackage{amsmath}  %
\usepackage{amssymb}  %
\usepackage{bm}       %

\theoremstyle{definition}

\renewcommand{\title}[1]{\newcommand{\titlelist}{{\huge\fontfamily{optimistic}\selectfont #1}}}

\newcommand{\paratitle}[1]{\vspace{1.5ex}\noindent\textbf{#1}}

\newcommand{\ignore}[1]{}

\usepackage{arydshln}
\definecolor{CQColor}{rgb}{0.0,0.0,1.0} %

\usepackage{graphicx}
\usepackage{colortbl}
\usepackage{amssymb}
\usepackage{pifont}
\usepackage{booktabs,multirow}
\usepackage{makecell}
\usepackage{tabulary}
\usepackage{fontawesome5}
\usepackage{bbding}
\usepackage{multicol}

\newlength\savewidth

\title{
\Large{\textcolor[HTML]{0369ff}{Ling and Ring 2.6} Technical Report} \\ [0.5em]
\huge{\textcolor[HTML]{0369ff}{Eff{}icient and Instant} Agentic Intelligence at Trillion-Parameter Scale}
}

\author[*]{Ling Team, Inclusion AI}

\contribution[*]{See Contributions section (Sec.~\ref{sec:contri}) for full author list.}

\abstract{
Efficient and scalable agentic intelligence requires models that can deliver both low-latency responses and strong reasoning capabilities while remaining practical to train, serve, and deploy. In this report, we present Ling-2.6 and Ring-2.6, a family of models designed to address this challenge at scale. Ling-2.6 is optimized for instant response generation and high capability per output token, whereas Ring-2.6 is tailored for deeper reasoning and more advanced agentic workflows.
Instead of training from scratch, we upgrade the Ling-2.0 base model through architectural migration pre-training and large-scale post-training. This upgrade is guided by a unified co-design of model architecture, optimization objectives, serving systems, and agent training environments, enabling improvements in both model capability and deployment efficiency.
At the architectural level, we introduce a hybrid linear attention design that integrates Lightning Attention with MLA, improving the efficiency of long-context training and decoding. To further enhance token efficiency, we optimize capability per output token through Evolutionary Chain-of-Thought, Linguistic Unit Policy Optimization, bidirectional preference alignment, and shortest-correct-response distillation.
For agentic capabilities, we propose KPop, a reinforcement learning framework designed to support stable training of Ring-2.6-1T on large-scale environment-grounded data. KPop improves training efficiency through asynchronous scheduling across coding, search, tool use, and workflow execution, enabling scalable learning from complex agent-environment interactions.
Together, Ling-2.6 and Ring-2.6 provide a practical pathway toward efficient, scalable, and open agentic systems. We open-source all checkpoints in the 2.6 family to support further research and development in practical agentic intelligence.
}

\date{June, 2026}
\gtechdata[Code]{\url{https://github.com/inclusionAI/Ling-V2.5}}
\gtechdata[Instant Model]{\url{https://huggingface.co/collections/inclusionAI/ling-26}}
\gtechdata[Thinking Model]{\url{https://huggingface.co/collections/inclusionAI/ring-26}}

\begin{document}
\maketitle

\input{sections/1-intro}
\clearpage
\input{sections/2-pre-training}

\clearpage

\input{sections/3-post-training}

\clearpage

\input{sections/4-infrastructure}

\clearpage

\input{sections/6-conclusion}

\input{sections/author}

\bibliographystyle{assets/plainnat}
\bibliography{main,ring}

\clearpage
\input{sections/8-appendix}

\end{document}

%% file: sections/1-intro.tex
\section{Introduction}

\begin{figure}[h!]
    \centering
    \begin{subfigure}[b]{1.0\textwidth}        
    \centering
        \includegraphics[width=0.9\linewidth]{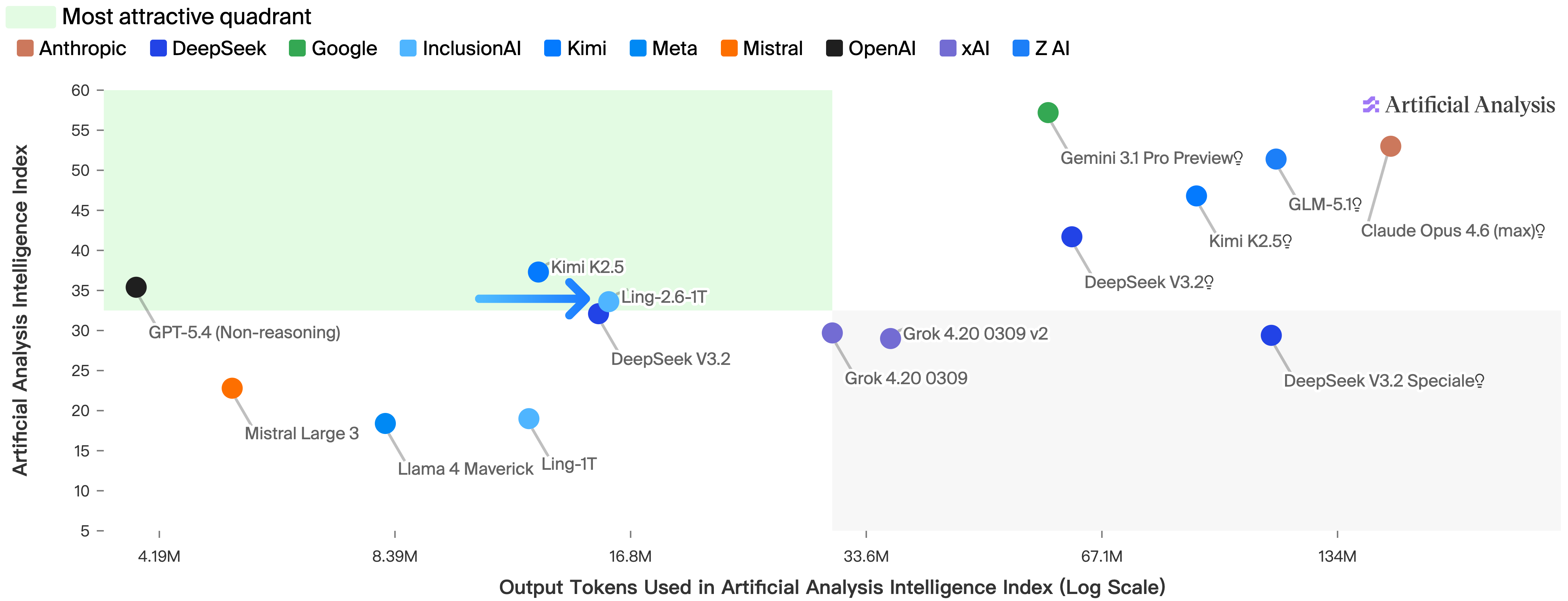}
      \end{subfigure}
    \vspace{2em}
    \begin{subfigure}[b]{1.0\textwidth}
    \centering
        \includegraphics[width=0.9\linewidth]{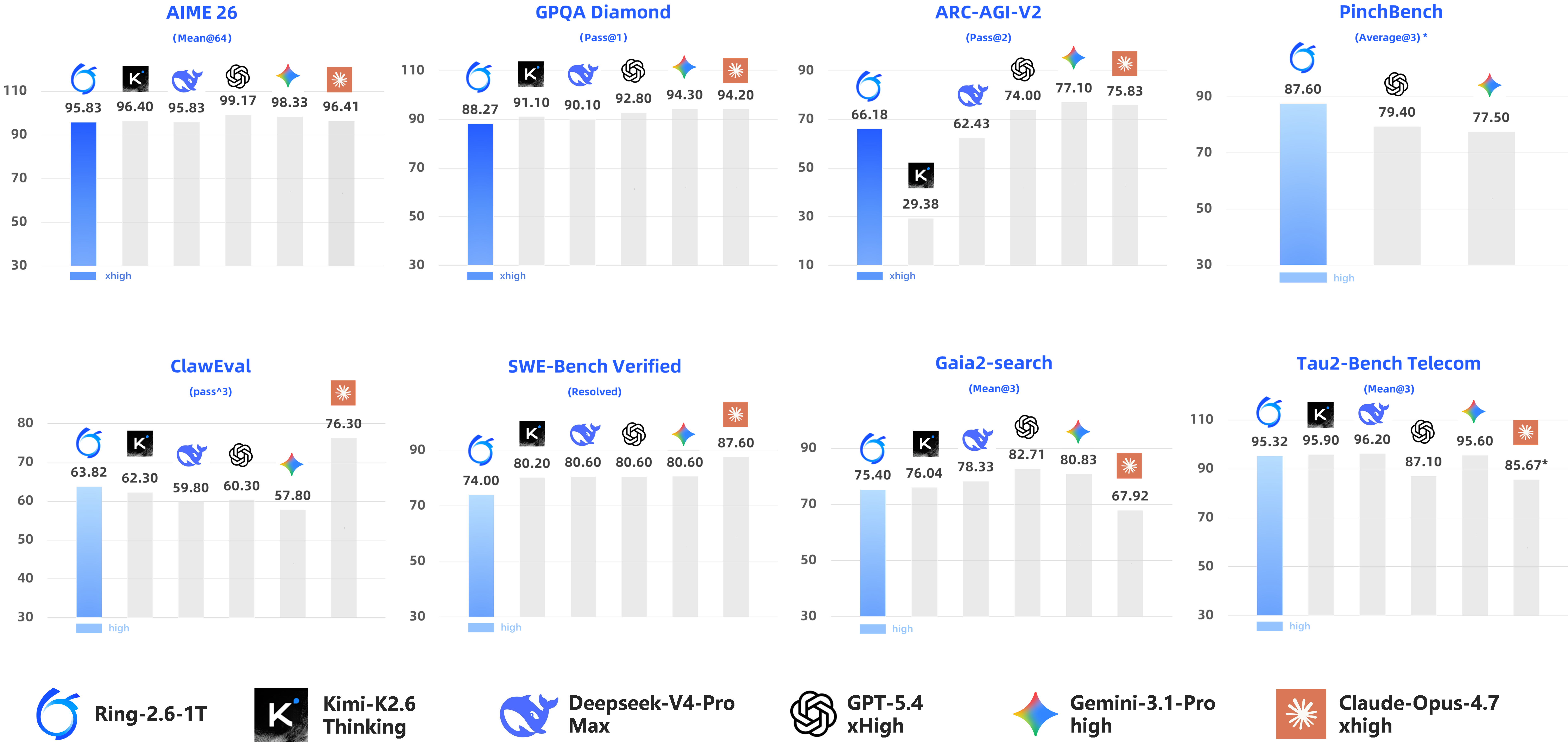}        
    \end{subfigure}
      
    \caption{\textbf{Up}: Ling-2.6-1T Artificial Analysis Intelligence Index: Score vs. Output Tokens Used. \textbf{Down}: Benchmark Performance of Ring-2.6-1T versus its counterparts. }
    \label{fig:intro_eval}
\end{figure}

Large language models (LLMs) are moving from chat systems to agentic systems~\citep{IntroducingClaudeOpus,IntroducingGPT552026,Gemini31Pro}.
This shift changes the optimization target of LLM. 
In other words, practical LLMs must reason well, use tools reliably, and stay efficient. 
However, these objectives are often in tension. 
Longer reasoning can improve performance, but it also raises latency and token cost. 
Fast models are easier to serve, but they often struggle on reliable reasoning and long horizon agentic tasks.
Thus, we argue that practical agentic intelligence should be improved along three directions at the same time: efficiency under long context, capability per output token, and native optimization for real agent workflows.

In this report, we introduce Ling-2.6 and Ring-2.6, a model family for efficient and instant agentic intelligence at trillion-parameter scale. 
In detail, Ling-2.6 is optimized for instant response and high token efficiency and Ring-2.6 is optimized for deeper reasoning with controllable effort. 
This model family scales from 104B to 1T parameters, with three open-sourced model weights: Ling-2.6-flash, Ling-2.6-1T, and Ring-2.6-1T. 
Rather than retraining a trillion-parameter model from scratch, we upgrade the Ling-2.0~\citep{teamEveryActivationBoosted2025} base model through architectural retrofit, continued pre-training, and large scale post-training. 
The following briefly summarizes the key design principles and results of the 2.6 family.

\paratitle{Efficient long context.}
Long contexts are valuable only when they are exploited with high efficiency.
In our earlier Group Query Attention (GQA)-based architecture~\citep{ainslie2023gqa,teamEveryActivationBoosted2025}, attention becomes the primary bottleneck as context length increases, exceeding 60\% of total FLOPs beyond 32K tokens. 
To overcome this limitation, Ling-2.6 and Ring-2.6 adopt a unified hybrid linear attention architecture that integrates Lightning Attention~\citep{qinTransNormerLLMFasterBetter2024} and MLA~\citep{deepseek-aiDeepSeekV2StrongEconomical2024} at a 7:1 ratio. 
This design preserves modeling quality while reducing long-context compute cost, KV-cache pressure, and decoding latency. 
At the same time, training a trillion-parameter model from scratch under a new architecture is prohibitively expensive.
We therefore continue pre-training from the Ling-2.0 base checkpoint for about 9.6T tokens, using a smooth architectural retrofit pipeline that includes hybrid initialization, QK Norm absorption, Partial RoPE adaptation, and MLA warmup. 
In parallel, we improve both serving and training through speculative decoding with continued MTP training, optimized context-parallel communication, and the linghe fused-kernel library. 
Together, these changes deliver higher decoding throughput than our GQA-based 2.0 generation.

\paratitle{High token efficiency.}
Token efficiency is a primary design target for Ling-2.6. 
We do not treat shorter responses as a superficial stylistic preference; rather, we optimize for higher capability per generated token. 
In post-training, we integrate Evolutionary Chain of Thought (Evo-CoT), Linguistic Unit Policy Optimization (LPO), and bidirectional preference alignment to increase information density while preserving reasoning fidelity~\citep{teamEveryActivationBoosted2025}. 
Evo-CoT removes redundant reasoning steps without relying on naive response-length minimization, while LPO shifts optimization from token-level actions to semantically coherent linguistic units, improving credit assignment and reducing wasteful repetition. 
Bidirectional preference alignment rewards informative, constraint-satisfying outputs and penalizes logical errors, hallucinations, and formulaic verbosity. 
We further apply shortest-correct-response distillation to increase capability per token. 
Collectively, these methods deliver approximately 4$\times$ higher token efficiency on reasoning workloads than the 2.0 generation. 
On the Artificial Analysis Intelligence Index, Ling-2.6-1T attains a score of 34 using only about 16M output tokens, comparable to GPT-5.4 in the non-reasoning setting.

\paratitle{Native agentic optimization.}
The agentic ability of the 2.6 family is trained directly rather than inherited indirectly from chat data. 
We construct broad agentic corpora spanning tool use, coding, search, workflow execution, and multi-turn interaction in real environments, and pair them with verifiable tasks, structured tool traces, and environment-grounded feedback. 
In addition, Ring-2.6 introduces KPop, a novel RL algorithm that replaces the uniform fixed-ratio constraint in IcePop~\citep{ring1t2025} with binary KL divergence to stabilize agentic reinforcement learning; combined with asynchronous RL that decouples rollout collection from parameter updates, this makes long, environment-bound trajectories tractable at trillion-parameter scale. 
This training recipe improves practical agent behavior across workflow and tool benchmarks. Ring-2.6-1T achieves 87.60 on PinchBench, 63.82 on ClawEval, and strong results on GAIA-2 Search and $\tau^2$-Bench Telecom. These results indicate that agent capability is a native training objective in the 2.6 family rather than a thin instruction-tuning layer.

Taken together, Ling-2.6 and Ring-2.6 push toward a more practical form of agentic intelligence. The architecture improves long-context efficiency. The post-training recipe improves intelligence per token. The agentic training stack improves reliability in real environments. 
Both base and post-training checkpoints of the 2.6 family are open-sourced and together with community efforts~\citep{teamQwenStudio2026,DeepSeek_V4pdfDeepseekaiDeepSeekV4Pro2026,KimiK26Tech,GLM51LongHorizonTasks} we are going to advance the frontier of efficient and open agentic intelligence.

%% file: sections/2-pre-training.tex
\section{Pre-training}
\label{sec:pretrain}

Long-horizon agentic tasks require efficient processing of ultra-long contexts.
At this scale, GQA dominates compute, yet training a trillion-parameter replacement from scratch discards the 20T-token investment in Ling-2.0-1T.
We instead retrofit the existing checkpoint with Lightning Attention and MLA in a 7:1 hybrid ratio, producing the shared base for both the Ling-2.5/2.6 and Ring-2.5/2.6 families.
Linear attention cuts per-token cost from $O(n^2)$ to $O(n)$; MLA compresses the KV cache into a low-rank latent space.
The question is whether this architectural transplant can succeed on an already-trained model and whether the resulting base remains viable for downstream specialization.
Section~\ref{sec:hybrid_attention} through Section~\ref{sec:eval_base_results} describe the architecture design, the four-stage migration strategy, the data and training recipe, and the evaluation results, exemplified on Ling-2.6-1T-base.

\begin{figure}[h!]
    \centering
    \includegraphics[width=0.75\textwidth]{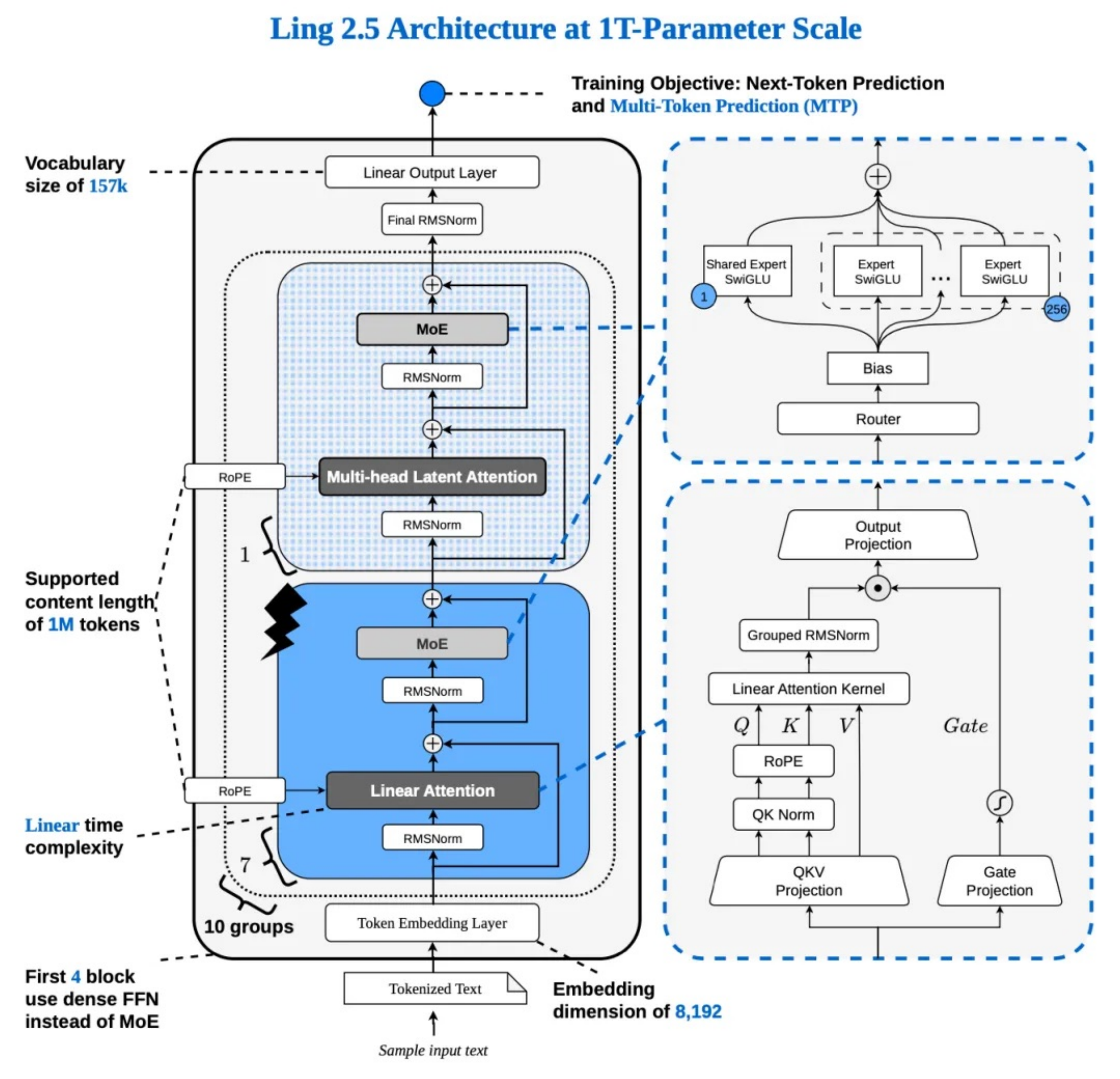}
    \caption{Overall architecture of Ling-2.6-1T-base. We employ a hybrid attention mechanism combining Lightning Attention and MLA in a 7:1 ratio, with fine-grained MoE for feed-forward layers.}
    \label{fig:architecture}
\end{figure}

\begin{table}[h!]
\centering
\caption{\textbf{Key architectural configurations of the Ling-2.6 series.}} %
\vspace{-0.1cm}
\label{tab:model-specs} %
\small
\begin{tabular}{l *{2}{>{\centering\arraybackslash}p{2.5cm}}}
\toprule
                        & \textbf{{Ling-2.6-flash}} & \textbf{{Ling-2.6-1T}} \\
\midrule
\# Layers                  & 32                 & 80              \\
\# Experts (total)         & 256                & 256             \\
\# Experts Active per Token& 8                  & 8               \\
\# Shared Experts          & 1                  & 1               \\
\# Attention Heads         & 32                 & 64              \\
\# Dense Layers  & 1                  & 4               \\
Hidden Size                & 4{,}096   & 8{,}192   \\
Intermediate Size          & 9{,}216   & 18{,}432  \\
Expert Intermediate Size  & 1{,}024   & 2{,}048    \\ 
KV LoRA Rank   & 512   & 512    \\ 
Q LoRA Rank   & 1536   & 1536    \\ 
Layer Group Size    & 8   & 8    \\ 
\bottomrule
\end{tabular}
\end{table}

\subsection{Hybrid Linear Attention Retrofit}
\label{sec:hybrid_attention}

Overall, Ling-2.6-1T-base incorporates three key architectural innovations to achieve efficient long-context modeling while preserving the pre-trained capabilities of the Ling-2.0-1T-base model.
First, we conduct scaling law experiments to determine the optimal hybrid ratio and validate an aggressive data switching strategy during continued training.
Then, based on these findings, we design a hybrid attention architecture combining Lightning Attention with MLA in a 7:1 ratio.
This substantially reduces inference FLOPs and KV cache memory in long-context scenarios.
Finally, we develop a smooth multi-stage migration strategy that enables performance-lossless conversion from the pre-trained Ling-2.0-1T-base checkpoint.
Figure~\ref{fig:architecture} illustrates the overall architecture, and the details are described below.

\subsubsection{Inherited Designs}
\label{sec:inherited}

\paratitle{Basic Configuration.}
We set the number of Transformer layers to 80 and the hidden dimension $d$ to 8{,}192. Each layer employs 64 attention heads with a head dimension of 128. The model uses a vocabulary size of 157{,}184 and supports a maximum context length of 262{,}144 tokens via RoPE~\citep{su2024roformer} positional encoding with $\theta = 6{,}000{,}000$. We employ SiLU as the activation function and RMSNorm with $\epsilon = 10^{-6}$ for layer normalization. The rotary dimension is set to 64, i.e., Partial RoPE is applied to a subset of head dimensions.

\paratitle{Mixture-of-Experts.}
Following Ling-2.0, Ling-2.6-1T-base employs a fine-grained MoE architecture for Feed-Forward Networks (FFNs), which sets routed experts and shared experts. Each MoE layer consists of 1 shared expert and 256 routed experts, where the intermediate hidden dimension of each routed expert is 2{,}048. Among the routed experts, 8 experts are activated for each token. We employ the auxiliary-loss-free load balancing strategy with expert bias enabled, where the router operates in FP32 precision with sigmoid scoring. We adopt a grouped routing strategy with $n_\text{group} = 8$ groups and top-4 group selection, and the routed output is scaled by a factor of 2.5 with normalized top-$k$ probabilities. For the first 4 Transformer blocks, we use dense FFN layers with an intermediate size of 18{,}432 rather than MoE layers.

\subsubsection{Hybrid Ratio Selection}
\label{sec:hybrid_ratio}

\begin{figure}[!hbt]
    \centering
    \includegraphics[width=0.5\textwidth]{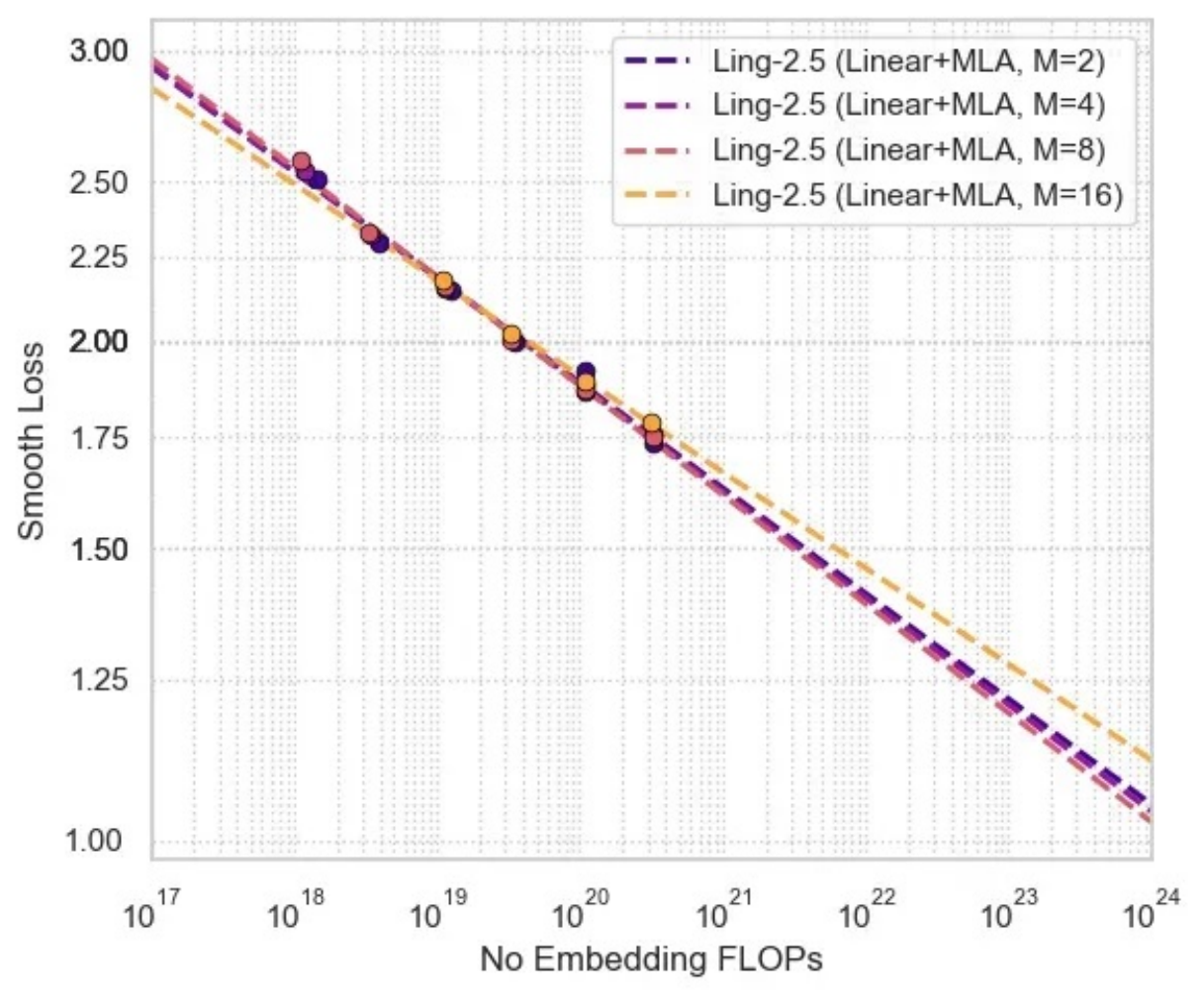}
    \caption{Scaling law curves for different hybrid ratios. Loss as a function of training FLOPs for $M = 2, 4, 8, 16$, where each group of $M$ layers contains 1 MLA layer and $M-1$ Lightning Attention layers.}
    \label{fig:scaling_law}
\end{figure}

We investigate the optimal ratio of Lightning Attention to MLA layers through scaling law experiments under equal FLOPs constraints. We define the layer group size $M$ such that each group contains 1 Full Attention (MLA) layer and $M-1$ Linear Attention layers, and compare configurations with $M \in \{2, 4, 8, 16\}$. Figure~\ref{fig:scaling_law} presents the scaling law curves, and Table~\ref{tab:hybrid_ratio} summarizes the results.

\begin{table}[t]
    \centering
    \caption{Comparison of hybrid ratio configurations under equal FLOPs constraints.}
    \label{tab:hybrid_ratio}
    \begin{tabular}{cccc}
        \toprule
        Layer Group Size ($M$) & Linear:Full Ratio & Scaling Performance & Inference Cost \\
        \midrule
        2  & 1:1  & Comparable to $M=4$    & High \\
        4  & 3:1  & Comparable to $M=2$    & Moderate \\
        \textbf{8}  & \textbf{7:1}  & \textbf{Best scaling trend} & \textbf{Low} \\
        16 & 15:1 & Notable loss degradation & Lowest \\
        \bottomrule
    \end{tabular}
\end{table}

The $M=8$ configuration (7:1 Linear-to-Full ratio) achieves the best scaling performance while providing substantial inference cost savings, representing the optimal balance between model quality and efficiency. We also observe that as the total FLOPs budget scales larger, the proportion of Linear Attention layers can be moderately increased without degradation. Based on these results, we select the 7:1 ratio for Ling-2.6-1T-base.

\subsubsection{Attention Transplantation}
\label{sec:attn_transplan}

As the context length increases, the attention mechanism emerges as the dominant computational bottleneck. We observe that at context windows exceeding 32K tokens, the computational cost shifts decisively from MoE layers to attention layers, and at 256K+ tokens, Full Attention (GQA) becomes the absolute performance bottleneck. To address this, we design a hybrid attention architecture that replaces a subset of GQA layers with Lightning Attention, while converting the remaining Full Attention layers from GQA to MLA for KV cache compression. This subsection describes the two attention components and the structural compatibility solutions required for MLA conversion.

\paratitle{Lightning Attention Conversion.}
\label{sec:lightning_attention}
We replace a portion of GQA layers with linear Lightning Attention layers, following the Ring-flash-linear-2.0~\citep{lingteam2025attentionmattersefficienthybrid}. During conversion, the GQA dimensions are expanded to standard Multi-Head Attention (MHA) by augmenting the $W_{qkv}$ projection along the head dimension. The newly introduced parameters are randomly initialized, and additional gating parameters $W_\text{gate}$ and gating normalization $\gamma_\text{gate}$ are introduced for the linear attention mechanism. During this initial conversion stage, we retain QK Norm and Partial RoPE from the original architecture to stabilize training, improve FP8 training compatibility, and enhance robustness at long-context windows.

\paratitle{MLA Conversion.}
\label{sec:gqa_to_mla}
MLA achieves extreme KV cache compression compared to GQA by projecting key-value pairs into a low-rank latent space. We validated this conversion through ablation experiments at two scales: a mini configuration (16B total, 1.3B activated) and a flash configuration (100B total, 5B activated). In both settings, replacing all GQA layers with MLA and training on 700B tokens followed by 600B mid-training tokens yielded performance that ultimately surpassed the original GQA baseline. This result held consistently across both the pure MLA and the hybrid Linear+MLA architectures.

However, directly converting Ling-2.0's attention to MLA raises two structural incompatibilities that must be resolved.

\paratitle{QK Norm Incompatibility.}
QK Norm is a nonlinear operation that prevents the KV weight matrix absorption required for efficient MLA inference. We resolve this by removing QK Norm prior to the MLA conversion. Leveraging the mathematical properties of RMSNorm, we approximately fuse the QK Norm parameters into the query and key projection weights. Specifically, for a dataset of $N$ calibration samples, we compute the per-dimension statistics of the query and key outputs $q_{ij}$ and $k_{ij}$ (for sample $i$ and dimension $j$ within head dimension $d$), and derive corrected projection weights $\hat{W}_q$ and $\hat{W}_k$ that absorb the effect of the QK Norm parameters $\gamma_Q$ and $\gamma_K$:

\begin{align}
    \hat{W}_q &\gets \frac{W_q}{\sqrt{\frac{1}{Nd}\sum\limits_{i=1}^{N}\sum\limits_{j=1}^d{q_{ij}^2} + \epsilon}} \odot \gamma_{Q}, \label{eq:qknorm_absorption_q} \\
    \hat{W}_k &\gets \frac{W_k}{\sqrt{\frac{1}{Nd}\sum\limits_{i=1}^{N}\sum\limits_{j=1}^d{k_{ij}^2} + \epsilon}} \odot \gamma_{K}. \label{eq:qknorm_absorption_k}
\end{align}

This calibration-based fusion effectively removes the nonlinear QK Norm while preserving its normalizing effect, enabling subsequent MLA weight absorption.

\paratitle{Positional Encoding Incompatibility.}
TransMLA~\citep{mengTransMLAMultiHeadLatent2025} natively supports Full RoPE, whereas Ling-2.0-1T-base employs Partial RoPE, in which only a subset of head dimensions receive rotary positional encoding. We address this by decoupling the RoPE module: the dimensions affected by RoPE are separated from unaffected dimensions, PCA-based weight rotation is applied exclusively to the RoPE-affected dimensions, and the results are concatenated to reconstruct the full representation. This decoupled treatment enables correct TransMLA conversion while preserving the Partial RoPE structure of Ling-2.0.

\subsection{Pre-training Corpus}

\subsubsection{Domain-specific Corpus}

\paratitle{Agentic Corpus.}
Agentic corpus represents a paradigm shift in pre-training data, moving beyond static text toward data that captures the complete information flow, decision pathways, and environmental dynamics an agent experiences during live deployment~\citep{zeng2026davinci}. The corpus features an exceptionally high diversity of tasks, spanning two primary domains: agentic tool use and agentic coding. Specifically, we construct a broad spectrum of tool-use tasks by leveraging over 500 real-world MCP environments encompassing more than 3,000 distinct tools; alongside this, we synthesize large-scale, diverse coding tasks integrated with bash commands, web-based QA, and relevant software repositories. Subsequently, we employ model-based quality filtering to ensure the relevance, logical coherence, and appropriate difficulty of the tasks. This is followed by the generation of agentic trajectories using diverse teacher models and various interaction scaffolds, incorporating rigorous rule-based and model-based verification to guarantee high-quality trajectories.

\paratitle{Long-Context Corpus.} 
To extend the context window to 256K and mitigate the scarcity of high-quality ultra-long corpora in naturally occurring data distributions, we jointly emphasized targeted retrieval and data synthesis, thereby constructing an ultra-long corpus covering multiple domains, including mathematics, complex web parsing, long-document summarization, retrieval-augmented generation (RAG) fusion, and multi-hop reasoning.

In parallel, we further enhanced the quality assurance pipeline. Specifically, by introducing a deep detection framework that integrates rule-based and model-based methods, we were able to effectively identify and remove common defects in ultra-long texts, including excessive self-repetition, structural collapse, and long-range semantic hallucinations. This pipeline was also applied to re-clean and refine a subset of the existing long-context data.

\subsubsection{General Corpus}

\paratitle{Web Corpus.}
To improve the general knowledge of model during pre-training, we built an efficient general feature engineering pipeline on top of a wide-table infrastructure and fastText~\citep{fasttext}, enabling hourly model iterations and daily feature updates. From a large-scale web index, we perform targeted STEM data recall, QA-oriented retrieval, and closed-loop retrieval via an internal search engine, and further rewrite the retrieved web content into textbook-style materials with reduced noise and stronger logical structure, effectively alleviating the knowledge coverage limitations of Common Crawl. To further improve the model factual question-answering capability, we observed that dispersed facts embedded in long-form text are difficult for models to learn effectively. To address this issue, we developed an atomic fact construction pipeline that converts Wikipedia articles into standalone propositions and structured triplets, together with a multi-strategy QA synthesis pipeline for fine-grained knowledge augmentation. Experimental results show that this approach substantially reduces the difficulty of factual memorization and effectively enhances the model factual question-answering ability.

\paratitle{Math \& Code Corpus.}
Mathematical and programming corpora remain the core components of our pre-training
data. For the current release, we have systematically augmented and refined these specific corpora via rephrasing corpus, and mining a broader spectrum of high-quality web content, comprehensive literature, and source code repositories.

\paratitle{Multilingual Corpus.}
To mitigate performance disparities across commonly-used languages in state-of-the-art multilingual models, we prioritize the expansion of monolingual corpora coverage for 21 languages, with focused additions in Arabic, Japanese, and Hindi. Regarding data composition, we incorporate 1.1T tokens from the open-source corpora Fineweb2~\citep{multilingual_fineweb2} and Fineweb2-hq~\citep{multilingual_fineweb2hq} (covering 8 key languages) and introduce approximately 70B tokens of synthetic web code data to enhance domain-specific representation capabilities. 

During the migration pre-training and continue pre-training, multilingual data receive a 4\% allocation of all datasets, internally stratified as 70\% web corpora—spanning Romance, Germanic, Slavic, and Southeast Asian language families—and 30\% capability-oriented mixes: mathematics , code, exams, synthetic web, and parallel corpora. And for mid-training, we strip away general web datasets, keeping only the high-value capability categories—mathematical reasoning, code, exams, synthetic data, and parallel corpora—to curate a low-noise, competency-focused distribution.

\subsection{Pre-training Recipe}
The pre-training of Ling-2.6 builds upon the Ling-2.0 language model checkpoint, inheriting its key hyper-parameters, and processes approximately 9.6T tokens across three stages.

\subsubsection{Hyper-Parameters. }
Ling-2.6-1T-base inherits the MoE backbone from Ling-2.0-1T-base and further introduces a hybrid linear attention architecture. Key architectural parameters scale with model size; see Table~\ref{tab:model-specs} for details.
For training hyper-parameters, we first re-establish the Ling scaling laws on the hybrid linear attention architecture to determine the appropriate learning rate and batch size given the total training budget. We adopt the auxiliary-loss-free load-balancing strategy, setting the bias-update rate $\gamma{=}0.001$ during continue pre-training, which is then reduced to 0.0001 for mid-training. The MTP loss weight is set to 0.1. All other hyper-parameters remain consistent with Ling-2.0. We continue to use the WSM scheduler from Ling-2.0: a linear warmup to a peak learning rate, followed by a constant phase until training concludes; the final annealing effect is achieved through checkpoint merging.

\subsubsection{Multi-Stage Training}

\begin{figure*}[t]
    \centering
    \includegraphics[width=0.95\textwidth]{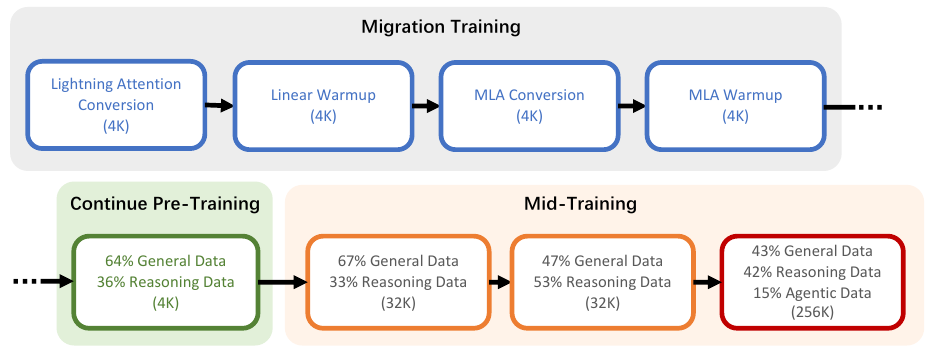}
    \caption{{Multi-stage pre-training pipeline of Ling-2.6.} The training spans approximately 9.6T tokens across three stages, progressively extending the context window from 4K to 256K.}
    \label{fig:multi_stage}
\end{figure*}

As illustrated in Figure~\ref{fig:multi_stage}, the pre-training processes approximately 9.6T tokens across three stages: Migration Pre-Training, Continue Pre-Training, and Mid-Training.

\paratitle{Migration Pre-Training.}\label{sec:migration}
Starting from the Ling-2.0 base checkpoint, this stage smoothly transitions the architecture from GQA-based Softmax Attention to the hybrid linear attention with MLA, while minimizing performance degradation. It proceeds in four steps over approximately 400B tokens.
First, we perform \textit{Lightning Attention Conversion}: a subset of GQA layers is converted to Lightning Attention by expanding $W_{qkv}$ parameters along the head dimension and initializing new gating parameters $W_\text{gate}$ and $\gamma_\text{gate}$, while retaining QK Norm and Partial RoPE for training stability.
Next, a \textit{Linear Warmup} phase freezes all parameters except $W_{qkv}$ and QK Norm weights. A learning rate warmup with a small data budget brings the loss back to pre-conversion levels, allowing the randomly initialized linear attention parameters to align with the pre-trained representations.
The third step is \textit{MLA Conversion}, which involves three sequential operations: (1) QK Norm removal via the calibration-based fusion described in Section~\ref{sec:gqa_to_mla}. At mini scale, this increases test perplexity from 6.65 to 11.13, a controlled and recoverable degradation; (2) a brief partial-parameter training phase (unfreezing only $W_{qkv}$, QK Norm, $W_\text{gate}$, and $\gamma_\text{gate}$) with learning rate warmup to mitigate the QK Norm removal impact; and (3) structural conversion via Partial RoPE adaptation and TransMLA~\citep{mengTransMLAMultiHeadLatent2025} conversion.
Finally, an \textit{MLA Warmup} freezes the parameters that were not structurally modified and applies a learning rate warmup to restore the loss to pre-conversion levels. At the end of Migration Pre-Training, the model has fully adopted the target hybrid linear attention architecture and is ready for large-scale continued training.

\paratitle{Continue Pre-Training.}\label{sec:continue-pretraining}
After migration pre-training, all parameters are unfrozen for full continued training on 8T tokens with a 4K context window. The data mixture allocates approximately 46\% to reasoning-intensive domains (e.g., mathematics and code), 50\% to general corpora (e.g., web text), and 4\% to multilingual data.
In preliminary experiments, we compare two data switching strategies: (1) a conservative approach that first trains on 2T tokens of the original pre-training data to recover model capabilities before introducing new, higher-quality data; and (2) an aggressive approach that introduces the new data from the outset of this stage. The aggressive strategy proves superior: early exposure to higher-quality data accelerates capability recovery, reduces the total tokens needed, and yields a higher final performance ceiling. We therefore adopt the aggressive strategy for Ling-2.6's Continue Pre-Training.

\paratitle{Mid-Training.}
In the mid-training stage, we train on high-quality data with long-context activation to refine capabilities and extend context windows. This stage comprises approximately 1.2T tokens and is divided into three phases.
The first phase trains on 250B tokens, sampling 20\% of sequences at 32K length while maintaining a data mixture similar to the previous stage, which expands the model's effective context window from 4K to 32K. The learning rate is kept at the peak value inherited from Continue Pre-Training.
The second phase spans 425B tokens at 32K context length, significantly increasing the proportion of high-quality data.
The third phase covers 525B tokens and extends the context window to 256K, while maintaining the high-quality data mixture.

\subsection{Pre-training Evaluation}

\label{sec:eval_base_results}

{\bfseries Benchmarks and Configurations.} 
To systematically evaluate the capabilities of the base model, we employ a broad benchmark suite spanning six key domains, including mathematics, coding, reasoning, language understanding, world knowledge, and long-context understanding. In total, the suite consists of 31 evaluation benchmarks, which are grouped into the following categories:
\begin{itemize}
    \item \textbf{Math}: GSM8K~\citep{gsm8k} (4-shot, CoT), CMath~\citep{cmath} (3-shot, CoT), 
    MathBench~\citep{mathbench} (4-shot, CoT), 
    OlympiadBench~\citep{olympiadbench} (3-shot, CoT), 
    OmniMath~\citep{omni-math} (3-shot, CoT), 
    GKMathUnion (4-shot, CoT). GKMathUnion is an in-house leaderboard composed of GaoKao 2023 En~\citep{gaokao2023_en} and GaoKao (including GaoKao I/II 2024\footnote{https://github.com/llmeval/Llmeval-Gaokao2024-Math}, GaoKao-Math-QA~\citep{agieval}, GaoKao-Math-Cloze~\citep{agieval} and 91 collected GaoKao problems in 2024).

    \item \textbf{Coding}: 
    HumanEval-Plus~\citep{evalplus} (0-shot), 
    HumanEval-Fim~\citep{humaneval_fim} (0-shot),  
    MBPP-Plus~\citep{mbpp+} (3-shot), 
    LiveCodeBench\footnote{LiveCodeBench contains 454 problems released between Aug 2024 and May 2025. }~\citep{livecodebench} (0-shot), 
    BIRD-SQL~\citep{birdsql} (0-shot).

    \item \textbf{General Reasoning}: 
    BBH~\citep{bbh} (3-shot, CoT), 
    BBH-zh~\citep{bbh_zh} (3-shot, CoT), 
    KorBench~\citep{korbench} (3-shot), CommonSenseQA~\citep{talmor2018commonsenseqa} (5-shot), WorldSense~\citep{worldsense} (0-shot), 
    AutoLogi~\citep{autologi} (3-shot, CoT), 
    ZebraLogic~\citep{zebralogic} (1-shot).

    \item \textbf{Language Understanding}: 
    Squad 2.0~\citep{squad2} (1-shot),
    Belebele~\citep{belebele} (0-shot).
    
    \item \textbf{World Knowledge}: 
    MMLU~\citep{mmlu} (5-shot), 
    MMLU-Pro~\citep{mmlu-pro} (5-shot), 
    GPQA~\citep{gpqa} (0-shot), 
    SuperGPQA~\citep{supergpqa} (5-shot), 
    TriviaQA~\citep{trivalqa} (5-shot),
    SimpleQA~\citep{simpleqa} (5-shot),
    C-SimpleQA~\citep{c-simpleqa} (5-shot),
    mARC~\citep{marc} (0-shot), 
    MMMLU\footnote{MMMLU language coverage may differ across baselines.}~\citep{mmmlu} (0-shot).
    
    \item \textbf{Long-Context Understanding}: 
    LEval~\citep{an2023leval} (0-shot),
    LongBenchv2~\citep{bai2025longbenchv2} (0-shot),
    
\end{itemize}
All evaluations are conducted within our internal evaluation framework, using identical benchmark-specific configurations to ensure a fair comparison among our four base models: Ling-2.0-flash-base, Ling-2.6-flash-base, Ling-2.0-1T-base, Ling-2.6-1T-base.

{\bfseries Evaluation Results.} 
Table~\ref{tab:ling-base-benchmarks} presents the evaluation results for the Ling-2.6-1T-base models and Ling-2.0-1T-base models. Overall, Ling-2.6-1T-base base models deliver broad and consistent gains over previous versions, with particularly notable improvements in knowledge-intensive evaluation, long-context modeling, and the preservation of strong mathematical and coding capabilities:
\begin{itemize}
    \item \textbf{Marked Gains in World Knowledge}:
    Both our Ling-2.6-flash-base and Ling-2.6-1T-base show substantial improvements in world-knowledge-oriented evaluations across English, Chinese, and multilingual benchmarks. Compared with Ling-2.0-1T-base models, they achieve notable gains on representative benchmarks such as GPQA, SimpleQA and MMMLU. These gains provide strong evidence that the high-quality knowledge data introduced during continued training has been effectively internalized by the models, leading to a stronger and more transferable knowledge representation.
    
    \item \textbf{Enhanced Long-Context and Reasoning Capabilities}: 
    Improvements are also evident in both long-context modeling and reasoning-intensive evaluations. Ling-2.6-1T-base base models can more effectively process extended inputs, preserve long-range dependencies, and identify task-relevant information from lengthy contexts, while exhibiting stronger multi-step reasoning and problem-solving abilities. These capabilities provide a stronger base-model foundation for enhancing agentic capabilities during post-training.

    \item \textbf{Robust Math and Code Performance under Continued Training}: 
    Despite additional training that substantially improves knowledge, long-context and reasoning capabilities, the models retain strong performance on mathematics and coding benchmarks, without noticeable degradation. In particular, performance further improves on representative benchmarks such as MathBench and LiveCodeBench, indicating that the training strategy effectively mitigates the common “seesaw effect” between knowledge expansion and reasoning-intensive capabilities.
\end{itemize}

\begin{table*}[htb]
\centering
\caption{Comparison among Ling-2.0-flash-base, Ling-2.6-flash-base, 
Ling-2.0-1T-base and Ling-2.6-1T-base.}
\label{tab:ling-base-benchmarks}
\resizebox{0.9\textwidth}{!}
{

\begin{tabular}{lcccc}
\toprule
\textbf{Benchmark} & \textbf{Ling-2.0-flash-base}  & 
\textbf{Ling-2.6-flash-base} & \textbf{Ling-2.0-1T-base} & 
\textbf{Ling-2.6-1T-base}\\
\midrule
\multicolumn{5}{c}{\textit{World Knowledge}} \\
\midrule
MMLU{\scriptsize(EM)} & 82.98 & \underline{84.13} & 86.03 & \textbf{86.82} \\
MMLU-Pro{\scriptsize(EM)} & 60.73 & \underline{61.36} & \textbf{67.91} & 67.79\\
GPQA{\scriptsize(EM)} & 35.35 & \underline{37.88} & 41.92 & \textbf{45.45}\\
SuperGPQA{\scriptsize(EM)} & 38.79 & \underline{40.17} & 44.06 & \textbf{44.72}\\
TriviaQA{\scriptsize(EM)} & 74.32 & \underline{75.52} & 81.62 & \textbf{82.62}\\
SimpleQA{\scriptsize(EM)} & 10.01 & \underline{18.33} & 20.87 & \textbf{38.26}\\
C-SimpleQA{\scriptsize(EM)} & 49.43 & \underline{63.53} & 64.53 & \textbf{76.83}\\
mARC{\scriptsize(EM)} & 82.07 & \underline{82.53} & 86.68 & \textbf{87.50}\\
MMMLU{\scriptsize(EM)} & 62.76 & \underline{64.76} & 68.68 & \textbf{71.53}\\
\midrule
\multicolumn{5}{c}{\textit{Math}} \\
\midrule
GSM8K{\scriptsize(Acc)} & 90.60 & \underline{91.89} & 89.31 & \textbf{93.93} \\
CMath{\scriptsize(Acc)} & 93.35 & \underline{93.53} & 93.62 & \textbf{94.72} \\
MathBench{\scriptsize(Acc)} & 77.69 & \underline{80.87} & 82.11 & \textbf{82.66} \\
GKMathUnion{\scriptsize(Acc)} & 63.17 & \underline{63.49} & 63.81 & \textbf{65.71} \\
OlympiadBench{\scriptsize(Acc)} & 35.70 & \underline{39.31} & 39.56 & \textbf{39.85} \\
OmniMath{\scriptsize(Acc)} & 28.30 & \underline{29.90} & 33.60 & \textbf{38.70} \\
\midrule
\multicolumn{5}{c}{\textit{Code}} \\
\midrule
HumanEval-Plus{\scriptsize(Pass@1)} & \underline{83.54} & 81.10 & 83.54 & \textbf{85.98}\\
HumanEval-Fim{\scriptsize(Pass@1)} & 80.93 & \underline{81.22} & 85.48 & \textbf{88.87} \\
MBPP-Plus{\scriptsize(Pass@1)} & \underline{74.07} & 73.28 & 73.81 & \textbf{77.78} \\
LiveCodeBench{\scriptsize(Pass@1)} & 30.40 & \underline{33.48} & 40.09 & \textbf{44.27} \\
BIRD-SQL{\scriptsize(Acc)} & \underline{38.69} & 38.40 & 42.70 & \textbf{44.59} \\

\midrule
\multicolumn{5}{c}{\textit{General Reasoning}} \\
\midrule
BBH{\scriptsize(Acc)} & 84.82 & \underline{85.06} & 86.88 & \textbf{89.73}\\
BBH-zh{\scriptsize(Acc)} & 83.59 & \underline{83.91} & 85.82 & \textbf{87.15} \\
KorBench{\scriptsize(Acc)} & 43.52 & \underline{44.96} & 49.04 & \textbf{50.64}\\
CommonSenseQA{\scriptsize(EM)} & \underline{87.71} & 87.31 & 89.76 & \textbf{90.99} \\
WorldSense{\scriptsize(EM)} & \underline{61.28} & 60.10 & 66.99 & \textbf{67.36} \\
AutoLogic{\scriptsize(Acc)} & 61.10 & \underline{62.82} & 65.76 & \textbf{67.43}\\
ZebraLogic{\scriptsize(Acc)} & 20.10 & \underline{21.00} & 26.00 & \textbf{30.00}\\
\midrule

\multicolumn{5}{c}{\textit{Language Understanding}} \\
\midrule
Squad 2.0{\scriptsize(Acc)} & \underline{89.27} & 88.81 & 91.29 & \textbf{92.98}\\
Belebele{\scriptsize(EM)} & 92.50 & \underline{93.22} & 93.89 & \textbf{94.67} \\
\midrule

\multicolumn{5}{c}{\textit{Long Context}} \\
\midrule
LEval{\scriptsize(Acc)} & 73.41 & \underline{77.86} & 72.30 & \textbf{76.21} \\
LongBenchv2{\scriptsize(Acc)} & 33.40 & \underline{34.19} & 30.02 & \textbf{43.54} \\
\bottomrule
\end{tabular}}
\end{table*}

%% file: sections/3-post-training.tex
\section{Post-training}
\label{section_post_training}
Post-training in Ling-2.6 and Ring-2.6 begins from a shared base model, but diverges in optimization target. 
Ling-2.6 is developed as an instant model, with emphasis on rapid response, high token efficiency, and basic agentic capability. 
Ring-2.6 follows the similar overall framework, while placing greater emphasis on stronger reasoning and more advanced agentic intelligence.

\subsection{Post-Training for Ling-2.6}
Unlike the unified post-training strategy employed in Ling-2.0, Ling-2.6 adopts an expert-driven training paradigm to systematically enhance capabilities across diverse domains. To this end, the Supervised Fine-Tuning (SFT) process is organized into two stages: an initial \textit{cold-start SFT} phase followed by \textit{specialized expert fine-tuning}. Reinforcement Learning (RL) is then applied to further strengthen each specialist model. Finally, the acquired domain-specific capabilities are distilled back into a unified Ling-2.6 model. This specialization-then-distillation framework is designed to maximize capability density while preserving the fast response characteristics and token efficiency required for practical deployment.
Additionally, we also adopt multiple MTP layer continued post-training for speculate decoding speedup (please refer to Appendix~\ref{sec:mtp_continue}).

\begin{figure}[h!]
    \centering
    \includegraphics[width=0.95\textwidth]{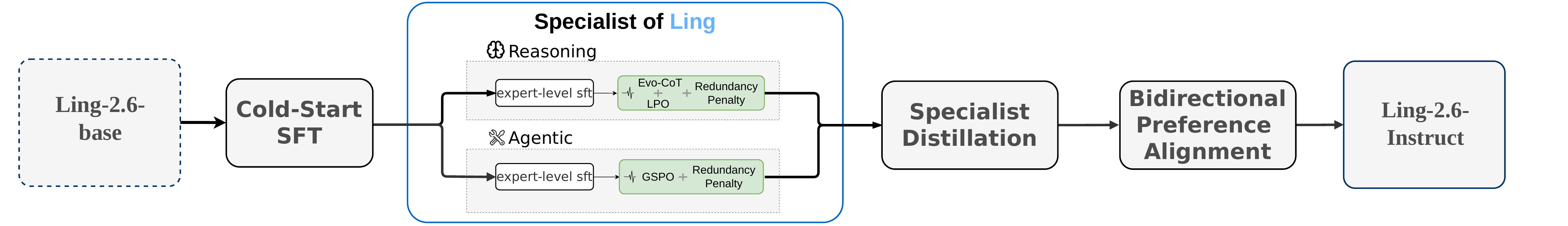}
    \caption{Post-Training Pipeline of Ling-2.6.}
    \label{fig:ling-pipline}
\end{figure}

\subsubsection{Supervised Fine-Tuning Corpus}
To provide a stable initialization for subsequent specialization in RL, the Ling-2.6 SFT corpus carefully balances reasoning, long-context window, and agentic tool use scenarios. This foundation is engineered to support token-efficient instant responses, deep long-context reasoning, and robust agentic behaviors.

\paratitle{Reasoning.} We compile a diverse reasoning dataset encompassing mathematics, STEM, coding, and logic. To maintain a balanced difficulty distribution, all samples are verified via model-based pass-rate annotations. We significantly expand the mathematical split with high-difficulty competition problems, broaden scientific coverage in chemistry and biology, and introduce over 50,000 synthesized logic queries across propositional and visual domains. Furthermore, we incorporate agent-centric coding tasks—such as repository exploration, debugging, and test generation—across multiple programming languages. Together, these verifiable, difficulty-controlled datasets provide optimal targets for reasoning-oriented RL.

\paratitle{Agent.}~~For agentic behavior and tool utilization, we construct RL data atop scalable, executable environments covering over 200 real-world and synthetic toolkits (e.g., search tools, Model Context Protocol servers). These environments offer more than 2,500 callable functions spanning search, e-commerce, finance, and scientific computation. Based on this infrastructure, we generate verifiable tool-use trajectories encompassing irrelevant tool rejection, long-horizon single-turn execution, and interactive multi-turn tasks~\citep{he2025vitabench}. By employing both seed-centric and critical-tool-chain-centric synthesis, the resulting corpus trains the model for accurate, concise, and highly efficient agentic interactions.

\paratitle{Long Context.} We extend the post-training context window to 256K tokens through a dedicated corpus comprising books, academic papers, code repositories, financial reports, and web data. To explicitly enhance long-context reasoning, we focus on number-dense corpora. We synthesize extended contexts by strategically merging multi-company and cross-year financial reports to create high-density, structured summaries. Instead of simple retrieval, we generate multi-hop calculation questions that force the model to reason across these extended contexts. To guarantee data quality, we employ a rigorous verification pipeline: an automated sandbox evaluates executable code for numerical accuracy, followed by human-in-the-loop validation, ensuring only high-quality, perfectly aligned reasoning traces are retained for training.

\subsubsection{Specialist Training of Ling}
Following the cold-start SFT, the pipeline shifts to a specialist training stage to develop individual expert models. This stage combines expert-level SFT with targeted Reinforcement Learning (RL). The core objective is to maximize instant-response quality under strict token-efficiency constraints, preparing these specialized capabilities for the final distillation into Ling-2.6.  

\paratitle{Reasoning.}
To instill adaptive reasoning capabilities while curbing verbosity, we implement a cohesive strategy spanning data refinement and reward design. We first refine the SFT reasoning data by utilizing proprietary expert models to generate responses, strictly retaining the shortest accurate candidate. To further eliminate "over-reflective" patterns—where redundant secondary reflection occurs after the correct answer has been found—we employ an LLM judge to prune these segments. This data-level intervention reduces the average output length by 200 to 300 tokens.

During the RL phase, we build upon the Evolutionary Chain of Thought (Evo-CoT) framework from Ling-2.0~\citep{teamEveryActivationBoosted2025}. Starting from the refined SFT checkpoint, we train the model using a composite reward system that explicitly penalizes redundancy:
\begin{itemize}
    \item \textbf{Accuracy} ($R_{\text{acc}}$): $+1$ if the final answer matches the ground truth; otherwise, $0$.
    \item \textbf{Formatting} ($R_{\text{format}}$): A penalty of $-0.5$ if explicit reasoning markers (e.g., ``\texttt{<think>}'' tags) are generated, enforcing an instant-response format.
    \item \textbf{Dynamic Length Penalty} ($\hat{R}_{\text{length}}$): Penalizes responses that exceed difficulty-specific length limits. It allows for elaborate reasoning on hard tasks while strictly curtailing length on easy tasks:
    \begin{equation}
        \hat{R}_{\text{length}} =
        \begin{cases}
            p(l), & \text{if } R_{\text{acc}} = 1, \\[4pt]
            \min\!\big(p(l),\, 0\big), & \text{if } R_{\text{acc}} = 0,
        \end{cases}
    \end{equation}
    where $p(l)$ is defined as:
    \begin{equation}
        p(l) = 0.5 - \frac{l - \ell_{\min}}{\ell_{\max} - \ell_{\min} + 10^{-9}}
    \end{equation}
    Here, $l$ denotes the token length of a sampled response, while $\ell_{\min}$ and $\ell_{\max}$ represent the shortest and longest lengths among all samples for a given query, respectively.
    \item \textbf{Semantic Redundancy Penalty} ($R_{\text{redundancy}}$): Relying solely on token length is insufficient to control repetitive internal reflection. Therefore, we segment the model's cognitive process and deploy an LLM judge to evaluate the semantic redundancy of each segment. Segments identified as logically circular or redundant are normalized into an additional penalty term $R_{\text{redundancy}}$, effectively forcing the model to think efficiently.
\end{itemize}

\paratitle{Agentic.}
For agentic tasks, we employ Group Sequence Policy Optimization (GSPO)~\citep{zheng2025group} tailored specifically to maximize token-efficient tool use. To achieve this, we introduce two dedicated reward signals. First, a \textit{process reward} measures the alignment between the model's predicted trajectory and the optimal ground-truth tool-call sequence~\citep{qian2026toolrl}, penalizing unnecessary or exploratory invocations to encourage concise execution. Second, we apply a \textit{compression-based repetition penalty} utilizing the zlib Compression Ratio. Since highly repetitive and degenerate outputs are highly compressible, they receive correspondingly harsher penalties, naturally fostering concise and coherent responses.

Beyond reward design, we improve training efficiency via a novel sample selection strategy called Dynamic Pass Rating (DPR). Instead of relying on static historical pass rates, DPR assesses task difficulty dynamically based on training behavior. From a temporal perspective, tasks solved early in training are deemed easy; from a stability perspective, tasks with consistently high pass rates are similarly classified. Conversely, tasks that remain unsolved or unstable are prioritized as hard. This creates an adaptive curriculum that constantly focuses training resources on informative samples near the model's current capability frontier.

\subsubsection{Bidirectional Preference Alignment}
As the final stage of efficiency-oriented post-training for Ling-2.6, we introduce a bidirectional focus reward mechanism. This aligns the model with fine-grained human preferences while actively preserving concise, information-dense responses. Unlike conventional unidirectional models, our design integrates positive incentives and negative penalties into a single reward model. This broadens the scoring range and provides preference gradients with a significantly higher signal-to-noise ratio. To prevent the model from hacking the reward by merely increasing output length, we design a focus reward mechanism~\citep{huang2026focalrewardbalancedreinforcement}. By monitoring on-policy saturation levels across different rubric dimensions, training weights are dynamically shifted away from saturated metrics toward dimensions requiring improvement. 

For complex tasks, we complement this general mechanism with specialized evaluators. For long-form writing, we propose ReportLogic framework~\citep{zhao2026reportlogicevaluatinglogicalquality}, which evaluates macro-structure and discourse organization, shifting the optimization objective from superficial fluency to deep textual logic. For multifaceted instruction following, an independent verification agent decomposes requests into explicit itemized rules for granular validation. While these task-specific rewards operate in their respective domains, the bidirectional focus reward remains the fundamental driver for general queries, ensuring that Ling-2.6 consistently delivers high-quality, token-efficient instant responses.

\subsection{Post-Training for Ring-2.6}

Ring-2.6 is further optimized for complex, long-horizon, and tool-intensive agentic behavior through specialist training, building on the post-training foundation of Ring-2.0~\citep{ring1t2025}.
The objective is not only to improve final-task success, but also to strengthen planning, search, tool use, and adaptive interaction under realistic execution constraints. 
Specifically, we construct a large-scale agentic post-training mixture spanning coding, search, and general tool-use tasks, pair it with reproducible execution environments, and develop an agentic reinforcement learning pipeline tailored to verifiable long-horizon behavior. Together, these components define the long-horizon agentic optimization stage.

After the training of specialists across both reasoning and agentic tasks, we apply specialist distillation to further enhance the overall performance. Moreover, we introduced adaptive thinking to the training pipeline to accommodate both daily uses and challenging reasoning problems. The adaptive thinking contains both SFT and RL stages. The $\textbf{high}$ mode employs moderate length penalties to balance reasoning depth with response conciseness, while the $\textbf{xhigh}$ mode uses minimal length penalties to maximize reasoning depth for complex reasoning problems.

\begin{figure}[h]
    \centering
    \includegraphics[width=0.95\textwidth]{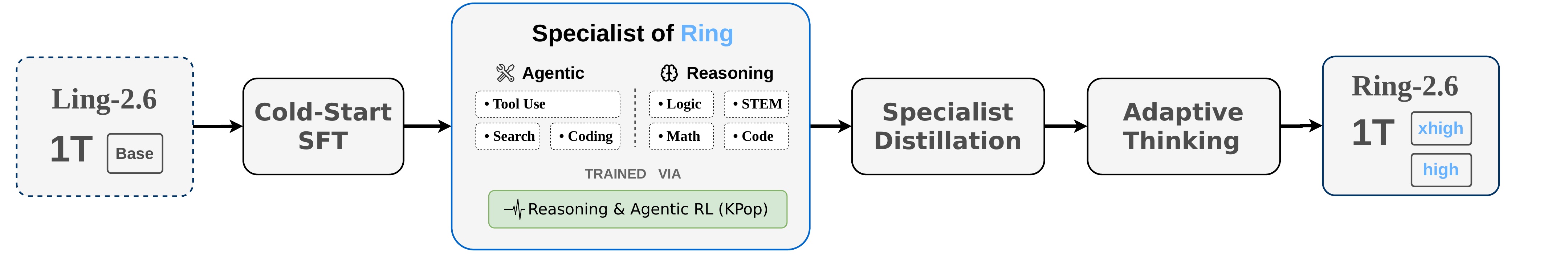}
    \caption{Post-Training Pipeline of Ring-2.6.}
    \label{fig:ring-train-pipeline}
\end{figure}

\subsubsection{Tool Use Data}
The tool-use data supports sustained agentic task execution under realistic constraints, emphasizing three complementary capabilities: repository-level coding, information seeking over mobile and web environments, and general-purpose workflows requiring planning and recovery from intermediate failures. Across all settings, we prioritize verifiability, environmental realism, and interaction diversity.

\paratitle{Coding Agent Tasks}

We mine PR-Issue pairs from GitHub at scale to train coding agents on real software engineering workflows. Starting from the full GH-Archive (Dec 2015--2023, $\sim$1B records), we retain only repositories with $>$100 stars, require merged PRs linked to closed issues, and mandate that each PR include a test patch for verifiability. Repositories overlapping with existing SWE benchmarks are excluded to prevent contamination. An LLM links PRs to their corresponding issues, yielding approximately 300K raw pairs. Combined with the reproducible execution environments described in Appendix~\ref{sec:agent_env}, these instances provide rigorous tasks for agentic reinforcement learning.

\paratitle{Search Agent Tasks}

\paragraph*{Mobile-App Search.}
Motivated by Meta ARE~\citep{froger2025are}, we construct synthetic personal digital universes with stateful applications (contacts, messages, emails, calendars, shopping, rides, apartments, files). Each universe maintains cross-app entity consistency with introduced near-miss entities and numeric fields for aggregation. Verifiable tasks are synthesized target-first: executable operation chains are paired with deterministic answers, then rewritten into natural requests. Rule-based and verifier-based filtering removes inconsistent or shortcut-solvable instances.

\paragraph*{Web Search.}
Starting from long-tail seed entities on the Wikipedia graph, we expand outward to accumulate constraints that are individually vague but jointly determine a unique answer, then rewrite them into natural questions using indirect references to resist lexical shortcuts. Instances answerable without retrieval are filtered out. Trajectories are collected with live tool execution, retaining only correct solutions and filtering out redundant tool calls and shortcut-style patterns.

Together, these two data sources strengthen information seeking under realistic interaction and context constraints, forming an important component of our agentic post-training mixture.

\paratitle{General-Purpose Agent Tasks}

Beyond coding and search, autonomous agents must handle diverse tool-use scenarios involving multi-step planning, policy compliance, and error recovery. We build general-purpose agent training data along four complementary directions:
\begin{itemize}
    \item \textbf{Policy-adherent multi-turn tool calling:} A five-stage pipeline synthesizes interactive tasks under business-policy constraints, covering boundary conditions and progressive difficulty. Reward signals are composed from environment state verification, tool-call sequence matching, and natural-language assertions, producing both SFT trajectories and RL task data.
    \item \textbf{Harness-agnostic workflow tasks:} Verifiable workflow tasks covering productivity, research, coding, data analysis, and document understanding are normalized around portable abstractions (user requests, tool schemas, observations, verification signals), enabling reuse across execution harnesses.
    \item \textbf{Large-scale MCP-based synthesis:} We synthesize multi-turn trajectories on 197 validated MCP servers spanning 12 domains with over 2{,}400 tools, requiring tool selection, error handling, and composition. Multi-layered quality control---including adversarial server detection and cross-batch deduplication---ensures data fidelity.
    \item \textbf{General tool-use tasks:} We expand to over 170 synthetic toolkits ($>$2{,}300 callable functions) across search, e-commerce, finance, and scientific computation. Tasks include irrelevant-tool, long-horizon single-turn, and interactive multi-turn settings~\citep{he2025vitabench}, each verified against environment state and critical tool-call inspection. Tasks are generated via seed-centric~\citep{fang2025towards} and critical-tool-chain paradigms.
\end{itemize}

To maintain training efficiency, we apply a Dynamic Pass Rating (DPR) strategy that selects tasks based on training dynamics: tasks mastered early or with consistently high pass rates are treated as easy, while persistently unsolved tasks are hard. This ensures the model trains on the most informative samples at its current capability frontier.

\medskip
Collectively, the coding, search, and general-purpose data described above form the core of our agentic post-training mixture. As reported in Section~\ref{sec:evaluation}, this data recipe yields competitive results on SWE-bench, GAIA2 Search, BrowseComp, $\tau^2$-bench, PinchBench, and ClawEval, confirming consistent improvements across coding, search, and general tool-use capabilities.

\subsubsection{Agentic Reinforcement Learning}
With the task mixture and execution environments in place, we train Ring-2.6 through an agentic reinforcement learning pipeline designed for verifiable long-horizon behavior. We develop a lightweight agent framework built on function-calling capabilities and equip the agent with three core tools: (1) \textbf{execute\_bash} for general bash command execution, (2) \textbf{search\_replace} for precise file editing, and (3) \textbf{task\_done} for signaling task completion. The maximum conversation length is set to 200 turns during training and 500 turns during evaluation.

We conduct reinforcement learning from a cold-start model in sandbox environments supported by AEnvironment~\footnote{https://github.com/inclusionAI/AEnvironment}. SWE tasks are inherently long-horizon and typically require 30 to 200 solution steps. To improve training efficiency, we apply several filtering stages to remove redundant and low-quality instances:
\begin{itemize}
    \item \textbf{Unstable instances}—where the gold patch occasionally fails the given tests—are removed.
    \item \textbf{Non-code edits}—instances whose gold patch involves changes outside source code (e.g., configuration or documentation only)—are excluded.
    \item \textbf{Complexity-based selection}—we compute the cold-start model's pass-rate on each instance as a complexity estimate and retain only those with pass rates in the 0.1–0.9 range that also carry well-defined issue descriptions annotated by human experts.
    \item \textbf{Per-repository deduplication}—to maintain diversity, we keep at most 3 instances per repository.
\end{itemize}
The resulting training set comprises approximately 2,500 instances drawn from 1,550 repositories spanning more than 30 programming languages, including Python, Java, C, Rust, and JavaScript.

To prevent the model from exploiting reward signals through information leakage, we adopt two safeguard mechanisms. First, under \textbf{Restricted Git History Access}, the `--all` flag is removed from `git log` commands, limiting the model's visibility to the commit history of the current branch. Second, under \textbf{Real-time Monitoring}, continuous monitoring is deployed during training to detect reward hacking behaviors. Our analysis reveals that approximately 0.2\% of trajectories exhibit cheating patterns, which remains negligible in practice.

\subsubsection{KPop: Bounding Mismatch with Binary KL Divergence}

In the release of Ring-1T~\citep{ring1t2025}, we proposed IcePop, which leverages double-sided masking to improve the training stability for reinforcement learning on MoE models. However, we observe that a uniform constant-ratio constraint implicitly assumes disproportional mismatch across tokens, which fails to reflect the heterogeneous training-inference discrepancy induced by different token probabilities. We therefore introduce KPop~\citep{KPop2026}, which replaces the uniform fixed-ratio constraint with binary KL divergence to better capture heterogeneous token-level mismatch across high- and low-probability regions. For the technical details, please refer to our blog\footnote{https://ringtech.notion.site/kpop}. 

Recall IcePop’s formulation, 
\begin{align}
\mathcal{J}_{\text{IcePop}}(\theta)
&=
\mathbb{E}_{
x \sim \mathcal{D},
\{y_i\}_{i=1}^{G}
\sim
\pi_{\textcolor{red}{\text{infer}}}
(\cdot \mid x; \theta_{\rm old})
}
\Bigg[
\frac{1}{G}
\sum_{i=1}^{G}
\frac{1}{|y_i|}
\sum_{t=1}^{|y_i|}
\Bigg[
\mathcal{M}
\Bigg(
\frac{
\pi_{\textcolor{blue}{\text{train}}}
(y_{i,t}\mid x,y_{i,<t};\theta_{\rm old})
}{
\pi_{\textcolor{red}{\text{infer}}}
(y_{i,t}\mid x,y_{i,<t};\theta_{\rm old})
},
\alpha,\beta
\Bigg)
\nonumber\\
&\qquad\qquad\qquad\qquad
\cdot
\min\Big(
r_{i,t}\widehat{A}_{i,t},
\text{clip}
\big(
r_{i,t},
1-\varepsilon,
1+\varepsilon
\big)
\widehat{A}_{i,t}
\Big)
\Bigg]
\Bigg].
\end{align}

IcePop adopts a uniform constant-ratio constraint on the policy probability ratio within a fixed global range $[\alpha, \beta]$, with additional double-sided masking, where a constant-ratio threshold treats all tokens the same way, regardless of their probability. But in reality, the noise in the ratio is not uniform across tokens, as the ratio divergence depends on token probability~\citep{KPop2026}. Thus, IcePop tends to over-mask the low-probability tokens.

KPop replaces IcePop's constant ratio bound with a symmetric binary KL criterion. For each output token $y_t,$ we compute the binary KL divergence between $\pi_{\textcolor{blue}{\text{train}}}(y_t)$ and $\pi_{\textcolor{red}{\text{infer}}}(y_t)$, which views the full vocabulary as a two-event partition, the current sampled token compared to everything else.

\begin{align*}
D_{\text{KL}}^B\!\left(
\pi_{\text{train}}(y_t)
\;\|\;
\pi_{\text{infer}}(y_t)
\right)
=
\pi_{\text{train}}(y_t)
\log
\frac{
\pi_{\text{train}}(y_t)
}{
\pi_{\text{infer}}(y_t)
}
+
\left(
1-\pi_{\text{train}}(y_t)
\right)
\log
\frac{
1-\pi_{\text{train}}(y_t)
}{
1-\pi_{\text{infer}}(y_t)
}
\end{align*}

An asymmetric variant checks only one direction, which is either forward or reverse. The symmetric KPop mask requires the binary KL to be small in \textit{both} directions.

\begin{align*}
\mathcal{M}_{\text{KPop}}(t) = \mathbf{1}\!\left[D_{\text{KL}}^B\!\left(\pi_{\textcolor{blue}{\text{train}}}(y_t) \| \pi_{\textcolor{red}{\text{infer}}}(y_t)\right) \leq \phi\right] \cdot \mathbf{1}\!\left[D_{\text{KL}}^B\!\left(\pi_{\textcolor{red}{\text{infer}}}(y_t) \| \pi_{\textcolor{blue}{\text{train}}}(y_t)\right) \leq \phi \right]
\end{align*}

The entire mechanism is controlled by a single hyperparameter $\phi$.

\begin{figure}[!hbt]
  \centering
  \includegraphics[width=\linewidth]{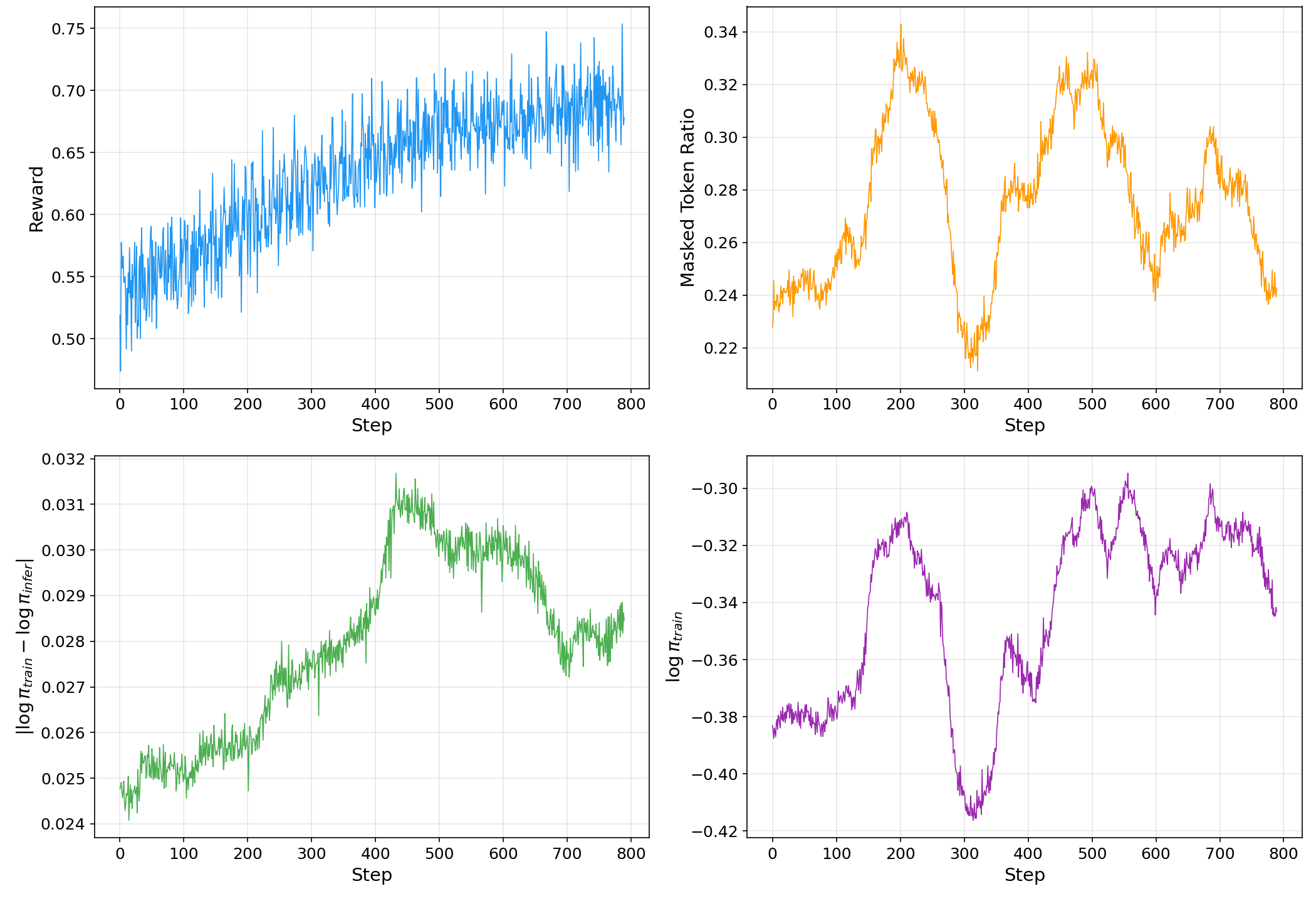}
  \caption{The training dynamics of agentic RL on coding task.}
  \label{fig:swe_curves}
\end{figure}

\begin{figure}[!hbt]
  \centering
  \includegraphics[width=0.6\linewidth]{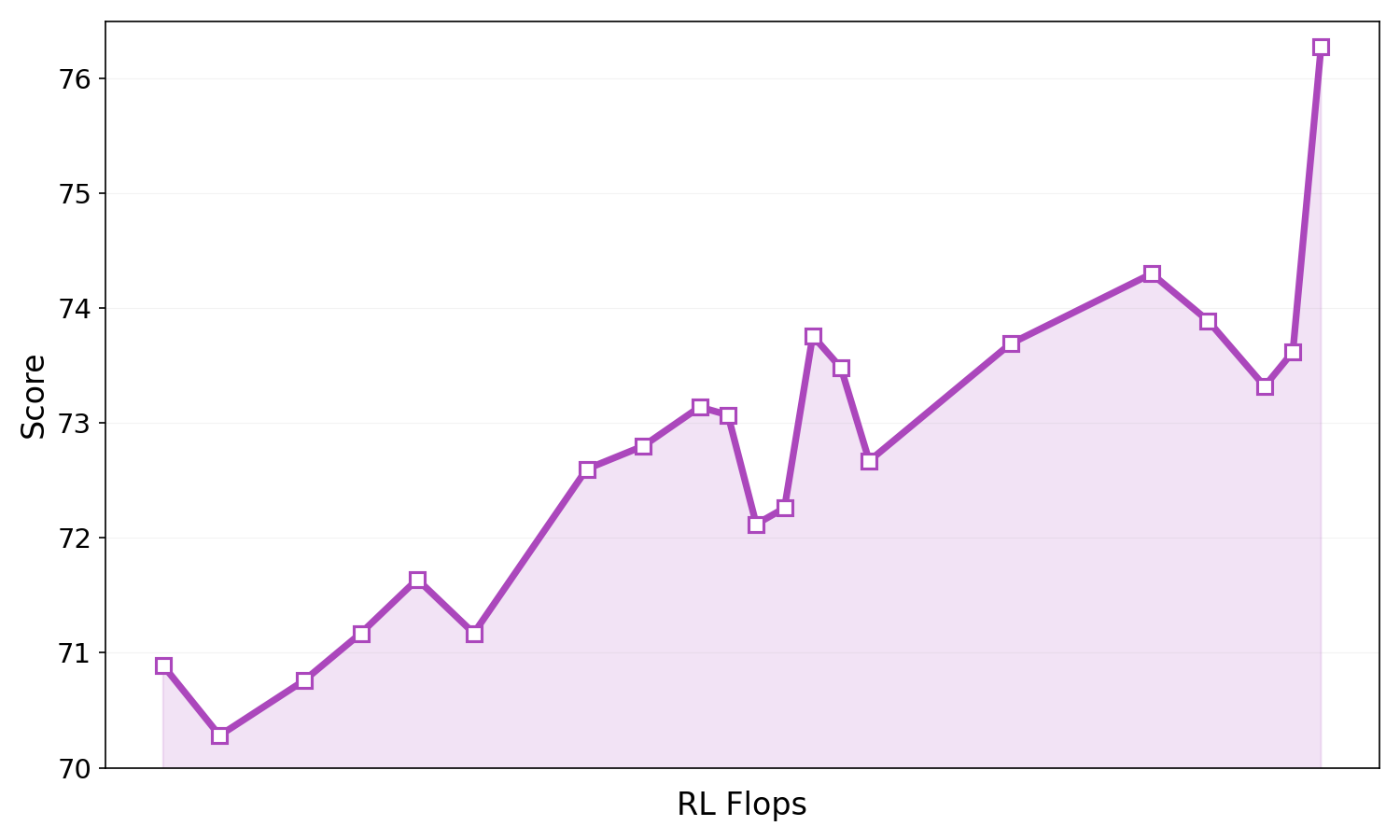}
  \caption{Evaluation on SWE-bench Verified.}
  \label{fig:swe_results}
\end{figure}

\paratitle{Empirical Results.}
Thanks to the help of Kpop, as shown in Figure~\ref{fig:swe_curves}, the reward curve exhibits a steady upward trajectory throughout training, rising from an initial value of 0.54 to approximately 0.68, which indicates consistent policy improvement. Figure~\ref{fig:swe_results} presents the evaluation results obtained during RL training. Notably, KPop enables effective agentic RL scaling on the lightweight agent, improving the solve rate on SWE-bench Verified from 70.8\% to 76.28\%, demonstrating the potential of KPop for large-scale agentic RL. All reported evaluation results are averaged over three independent runs.

\subsection{Evaluation}
\label{sec:evaluation}

In this section, we present a comprehensive evaluation of Ling-2.6 and Ring-2.6 across a broad set of benchmarks covering knowledge, reasoning, agentic behavior, instruction following, and long-context understanding. The evaluation is designed to assess not only overall capability, but also the distinct optimization targets of the two model families. In particular, Ling-2.6 is examined as an instant model optimized for fast response and token efficiency, while Ring-2.6 is evaluated as a stronger long-horizon agentic model with enhanced reasoning and tool-use capabilities. Together, these results provide a systematic view of the capability profile and specialization of our flagship models.

\subsubsection{Evaluation of Ling-2.6}
We present a comprehensive evaluation of the {Ling-2.6} series models, specifically the high-performance {Ling-2.6-1T} and its lightweight counterpart, {Ling-2.6-flash}. Their capabilities are assessed across a diverse array of benchmarks spanning five key domains: knowledge, reasoning, agentic capabilities, instruction following, and long-context understanding. We compare their performance against a wide range of state-of-the-art systems, including leading proprietary models in their non-reasoning or instant modes (e.g., {Kimi-K2.5}, {GPT-5.4}, {GLM-5}, {Nemotron-3-Super}, {DeepSeek-V3.2}). 

\paratitle{Benchmarks}

To comprehensively assess the Ling-2.6 series models, we conduct evaluations across a wide range of benchmarks, primarily covering 5 domains: knowledge, reasoning, agentic capabilities, instruction following, and long-context understanding.

\begin{itemize}
    \item \textbf{Knowledge}: C-SimpleQA~\citep{c-simpleqa} (Correct), SimpleQA-Verified~\citep{simpleqa} (Correct), GPQA-Diamond~\citep{rein2023gpqa} (Mean@4, CoT), SuperGPQA~\citep{supergpqa} (EM, CoT), and Humanities-Last-Exam~\citep{hle} (Mean@4).
    \item \textbf{Reasoning}: AIME 2026\footnote{\url{https://artofproblemsolving.com/wiki/index.php/AIME_Problems_and_Solutions}} (Mean@64, CoT), HMMT (Nov25, Feb26)~\citep{HMMT25} (Mean@64, CoT), IMO-AnswerBench~\citep{luong-etal-2025-towards} (Mean@8, CoT), LiveCodeBench-v6~\citep{jain2025livecodebench} (Mean@4), bbeh~\citep{bbeh} (Pass@1), and ARCPrize~\citep{chollet2024arcprize} (Mean@4).
    \item \textbf{Agentic}: SWE-bench Verified~\citep{jimenez2024swebench} (Claude Code, openhands fc), PinchBench\footnote{\url{https://pinchbench.com/}} (Mean@5), ClawEval\footnote{ClawEval is the evaluation suite for OpenClaw agents; see \url{https://kimi.com/blog/kimi-k2-6} for benchmark details.} (Pass@3), BFCL-V4~\citep{berkeley-function-calling-leaderboard} (Accuracy), $\tau^2$-bench~\citep{barres2025tau2bench} (Mean@4, user model gpt-5.2, include Average, Retail, Airline, Telecom), and terminal-bench 2.0~\citep{merrill2026terminal} (Accuracy).
    \item \textbf{Instruction Following}: IFBench~\citep{pyatkin2025generalizing} (Mean@5) , LIFEBench~\citep{zhang2026lifebench}.
    \item \textbf{Long-Context \& Dialogue}: LongBenchv2~\citep{bai2025longbenchv2} (Accuracy), MRCR~\citep{vodrahalli2024michelangelo} (Mean@16K-256K), Multichallenge~\citep{deshpande-etal-2025-multichallenge} (Accuracy), and Multi-IF~\citep{he2024multi} (turn-3).
\end{itemize}

We evaluate our models, {Ling-2.6-1T} and {Ling-2.6-flash}, against a diverse set of leading and open-source models. The comparison includes various non-reasoning or instant-response configurations of proprietary models, such as Kimi-K2.5, DeepSeek-V3.2, GPT-5.4, GPT-5.4-mini, GLM-5, GLM-4.5-Air, Nemotron-3-Super-120B-A12B and GPT-OSS-120B.

\begin{table*}[htb]
  \centering
  \caption{Comparison of Ling-2.6-flash and other models on various Benchmarks.}
  \label{tab:ling-2.6-flash_results}
  \resizebox{0.9\textwidth}{!}{
  \begin{tabular}{lcccc}
    \toprule
    \textbf{Benchmark} & \textbf{\splitcell{{Ling-2.6-flash}}} & \textbf{\splitcell{{Nemotron-3-Super}\\{120B-A12B}\\{non-reasoning}}} & \textbf{\splitcell{{GPT-OSS-120B}\\{low}}} & \textbf{\splitcell{{GPT-5.4-mini}\\{non-reasoning}}} \\
    \midrule
    \multicolumn{5}{c}{\itshape Knowledge} \\
    \midrule
    C-SimpleQA             & \textbf{60.23} & 47.03 & 43.47 & 59.73 \\
    SimpleQA-Verified      & 15.10 & \textbf{17.1} & 12.1 & 16.5 \\
    Humanities-Last-Exam   & 6.30 & \textbf{11.26} & 4.59 & 5.61 \\
    \midrule
    \multicolumn{5}{c}{\itshape Reasoning} \\
    \midrule
    AIME26                 & 73.85 & \textbf{88.59} & 60.10 & 35.68 \\
    HMMT-Feb26             & 49.29 & \textbf{76.23} & 35.89 & 21.73 \\
    IMO-AnswerBench        & 54.28 & \textbf{79.53} & 38.59 & 21.16 \\
    LiveCodeBench-v6 & 62.28 & \textbf{74.67} & 61.51 & 55.07 \\
    \midrule
    \multicolumn{5}{c}{\itshape Agentic} \\
    \midrule
    SWE-bench-Verified     & \textbf{61.20} & 61.00 & --    & 43.80 \\
    PinchBench             & \textbf{81.30} & 73.10 & 52.00 & 71.40 \\
    BFCL-v4                & \textbf{66.81} & 35.12 & 43.30 & 49.72 \\
    $\tau^2$-bench             & \textbf{76.36} & 68.92 & 23.48 & 46.92 \\
    ClawEval              & \textbf{64.56} & 56.00 & --    & --    \\
    \midrule
    \multicolumn{5}{c}{\itshape Instruction Following} \\
    \midrule
    IFBench                & 57.40 & 39.87 & \textbf{58.30} & 38.80 \\
    LIFEBench              & 57.20 & 44.60 & 49.40 & \textbf{57.60} \\
    \midrule
    \multicolumn{5}{c}{\itshape Long-context \& Multi-turn Dialogue} \\
    \midrule
    Multichallenge         & \textbf{39.71} & 35.16 & 37.73 & 34.43 \\
    Multi-IF (turn-3)      & \textbf{74.80} & 64.80 & 68.52 & 71.13 \\
    MRCR (16K-256K)        & \textbf{75.93} & 39.04 & 22.56 & 34.76 \\
    \bottomrule
  \end{tabular}
    }
\end{table*}

\begin{table*}[htb]
  \centering
  \caption{Comparison of Ling-2.6-1T and other state-of-the-art models on Different Benchmarks.}
  \label{tab:Ling-2.6-1T_results_revised}
  \small
  \begin{tabular}{lccccc}
    \toprule
    \textbf{Benchmark} & \textbf{\splitcell{{Ling-2.6-1T}}} & \textbf{\splitcell{{GLM-5}\\{non-thinking}}} & \textbf{\splitcell{{GPT-5.4}\\{non-reasoning}}} & \textbf{\splitcell{{DeepSeek-V3.2}\\{nothink}}} & \textbf{\splitcell{{Kimi-K2.5}\\{Instant}}} \\
    \midrule
    \multicolumn{6}{c}{\itshape Knowledge} \\
    \midrule
    C-SimpleQA             & 76.53 & 71.97 & 70.57 & 68.37 & \textbf{76.80} \\
    SimpleQA-Verified      & \textbf{31.50} & 28.90 & 30.20 & 23.70 & 25.40 \\
    GPQA-Diamond           & 76.17 & 70.20 & 76.89 & 77.11 & \textbf{80.52} \\
    SuperGPQA              & 58.32 & 57.63 & 62.97 & 61.37 & \textbf{66.40} \\
    Humanities-Last-Exam   & 10.06 & 7.09 & 10.75 & 10.47 & \textbf{12.92} \\
    \midrule
    \multicolumn{6}{c}{\itshape Reasoning} \\
    \midrule
    LiveCodeBench-v6 & 65.58 & 52.09 & 70.76 & 57.71 & \textbf{73.40} \\
    bbeh                   & \textbf{52.37} & 27.80 & 25.70 & 48.04 & 48.43 \\
    AIME26                 & \textbf{87.40} & 49.22 & 72.92 & 66.41 & 66.98 \\
    HMMT-Nov25             & \textbf{81.93} & 46.09 & 47.76 & 53.44 & 61.20 \\
    IMO-AnswerBench        & \textbf{65.81} & 39.34 & 44.75 & 46.66 & 52.56 \\
    ARCPrize               & \textbf{50.94} & 12.31 & 26.00 & 20.06 & 31.19 \\
    \midrule
    \multicolumn{6}{c}{\itshape Agentic} \\
    \midrule
    SWE-bench-Verified     & 72.20 & \textbf{73.80} & 69.20 & 66.40 & 66.80 \\
    PinchBench             & 85.24 & 83.29 & 73.40 & 85.38 & \textbf{85.48} \\
    ClawEval              & \textbf{51.00} & 40.38 & 43.26 & 46.15 & 48.08 \\
    BFCL-v4                & \textbf{70.64} & 67.57 & 56.09 & 60.05 & 62.96 \\
    $\tau^2$-bench             & \textbf{78.36} & 78.12 & 69.53 & 75.63 & 71.21 \\
    terminal-bench 2.0     & 40.45 & \textbf{48.31} & 46.07 & 29.21 & 48.30 \\
    \midrule
    \multicolumn{6}{c}{\itshape Instruction Following} \\
    \midrule
    IFBench                & \textbf{57.62} & 57.14 & 49.73 & 50.00 & 42.53 \\
    \midrule
    \multicolumn{6}{c}{\itshape LongText} \\
    \midrule
    LongBenchV2            & 48.31 & /     & 49.30 & 51.89 & \textbf{59.64} \\
    MRCR(16K-256K)         & \textbf{80.37} & /     & 68.43 & 30.50 & 63.22 \\
    \bottomrule
  \end{tabular}
\end{table*}

\paratitle{Results}

The evaluation results, presented in Table~\ref{tab:ling-2.6-flash_results} and Table~\ref{tab:Ling-2.6-1T_results_revised}, demonstrate the strong and versatile capabilities of the {Ling-2.6} series models across multiple domains.

\paragraph*{Knowledge.}
In knowledge-based benchmarks, both models exhibit strong performance. {Ling-2.6-1T} achieves a top-tier score of 76.53 on {C-SimpleQA} and a dominant 31.50 on {SimpleQA-Verified}, outperforming most competitors. While {Kimi-K2.5} shows an edge on highly specialized exams like {GPQA-Diamond} (80.52) and {Humanities-Last-Exam} (12.92), {Ling-2.6-1T} remains highly competitive. Meanwhile, the lightweight {Ling-2.6-flash} leads its comparison group with a score of 60.23 on {C-SimpleQA}, surpassing other fast models like {GPT-5.4-mini} (59.73).

\paragraph*{Reasoning.}
The reasoning domain highlights the exceptional strength of {Ling-2.6-1T}. It establishes clear leadership by achieving top scores across a majority of challenging reasoning benchmarks, including {AIME26} (87.40), {HMMT-Nov25} (81.93), {IMO-AnswerBench} (65.81), {bbeh} (52.37), and {ARCPrize} (50.94). This performance significantly surpasses other non-thinking/instant models. In contrast, while {Ling-2.6-flash} provides solid baseline reasoning capabilities (e.g., 73.85 on {AIME26}), it is outperformed by larger open-weight models like {Nemotron 3 Super}, underscoring the performance-efficiency trade-off for this lightweight model.

\paragraph*{Agentic Capabilities.}
On agentic tasks, both models demonstrate robust and often state-of-the-art performance. {Ling-2.6-1T} shows its versatility by leading on {PinchBench} (85.24), {ClawEval} (51.00), {BFCL-v4} (70.64), and {$\tau^2$-bench} (78.36), and remains highly competitive on software engineering ({SWE-bench-Verified}) and terminal operation tasks. More impressively, {Ling-2.6-flash} proves to be a standout performer in this category. Despite being a lightweight model, it consistently leads its comparison group on complex tasks like {SWE-bench Verified} (61.20), {PinchBench} (81.30), and {$\tau^2$-bench} (76.36), indicating a remarkable aptitude for tool use and task execution.

\paragraph*{Instruction Following and Long-Context.}
The models' abilities to follow instructions and process long texts are excellent. {Ling-2.6-1T} achieves a leading score of 57.62 on {IFBench} and a dominant 80.37 on {MRCR (16K-256K)}, showcasing superior long-context retrieval and understanding. {Ling-2.6-flash} is particularly exceptional in this area, decisively outperforming its peers across all listed long-context and dialogue benchmarks. It scores 75.93 on {MRCR}, more than doubling the score of its closest competitors, and also leads on {Multichallenge} (39.71) and {Multi-IF} (74.80), confirming its outstanding efficiency and effectiveness in handling long and complex interactions.

\paragraph*{Token Efficiency.} We conduct Artificial Analysis Intelligence Index evaluation on {Ling-2.6-1T}, as shown in the upper part of Figure~\ref{fig:intro_eval}. The model achieves a score of 34 using only about 16M output tokens, which is about 4$\times$ better token efficiency than Ling-2.0-1T and comparable to GPT-5.4 in the non-reasoning setting. This highlights the significant improvement in token efficiency achieved through our post-training recipe, enabling high-quality reasoning with substantially fewer output tokens.

\begin{figure}[htb]
    \centering
    \begin{subfigure}[b]{1.0\textwidth}        
    \centering
        \includegraphics[width=0.47\linewidth]{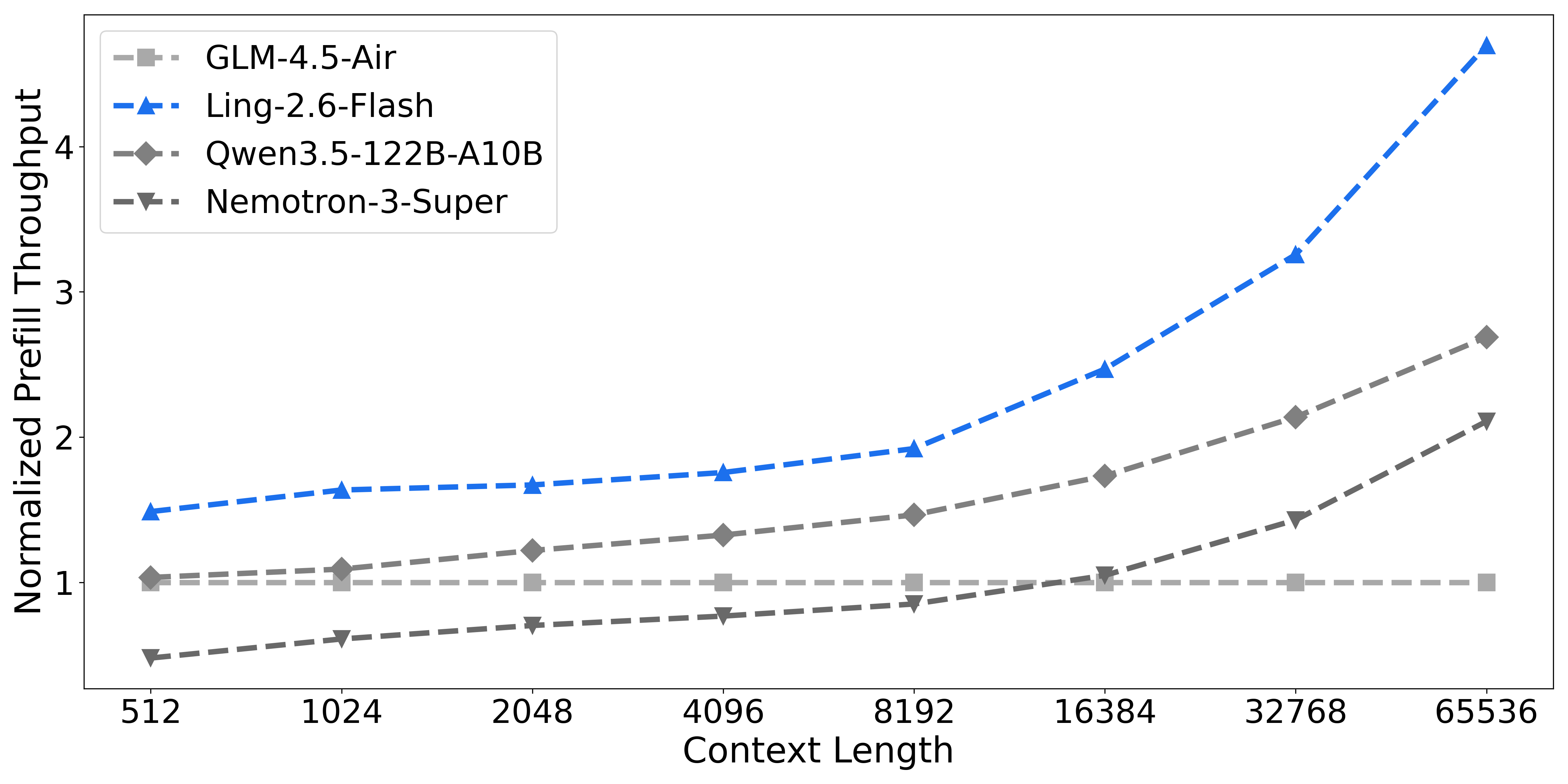}
        \hfill
        \includegraphics[width=0.47\linewidth]{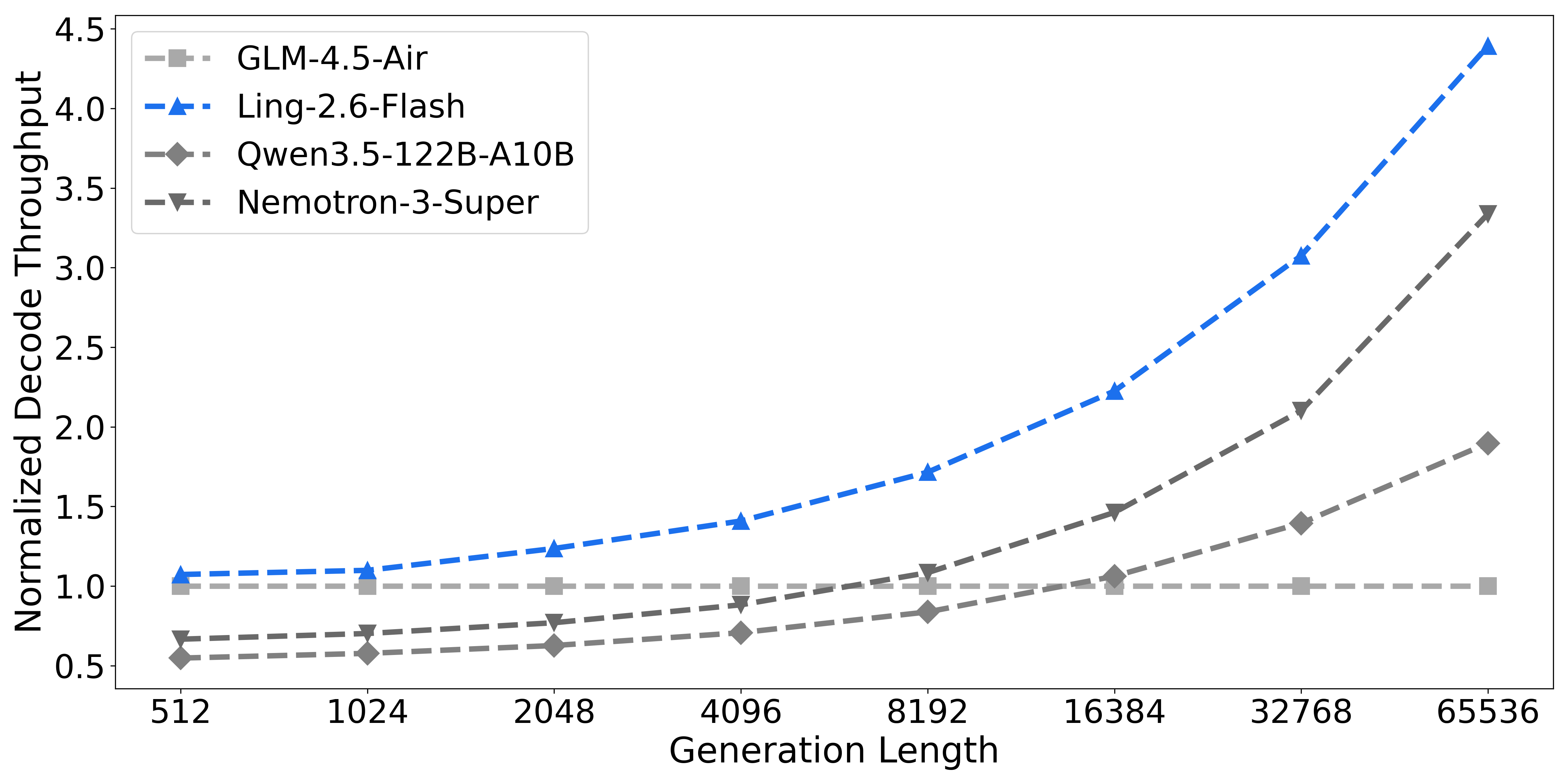}        
      \end{subfigure}
    \caption{The prefill and decode throughput performance of Ling-2.6-flash.}
    \label{fig:ling-v26-infer-effi}
\end{figure}

\paragraph*{Inference Efficiency.} Beyond benchmark quality, {Ling-2.6-1T} primarily targets the highest capability regime and therefore carries substantially higher deployment requirements, whereas {Ling-2.6-flash} is explicitly designed for deployment-time inference efficiency.
Benefiting from the hybrid attention design and a highly sparse MoE architecture, {Ling-2.6-flash} delivers substantially faster serving than state-of-the-art models in a similar size regime, with up to 4$\times$ acceleration in both prefill and decode. 
As shown in Fig~\ref{fig:ling-v26-infer-effi}, under a 4$\times$H20 deployment, batch size 32, 4 tensor parallel ranks, and output length 64K, the decode throughput of {Ling-2.6-flash} is 1.3$\times$ that of Nemotron-3-Super, 2.4$\times$ that of Qwen3.5-122B-A10B, and 4.3$\times$ that of GLM-4.5-Air. 
In practice, this efficiency profile makes Ling-2.6-flash better suited to latency-sensitive, long-output, and throughput-constrained agentic workloads, where response speed is part of model utility rather than a secondary systems consideration.

\subsubsection{Evaluation of Ring-2.6}
In this section, we present a comprehensive evaluation of {Ring-2.6-1T} across a wide range of benchmarks spanning reasoning, OpenClaw, agentic coding, agentic search, and function calling. We compare its performance against leading frontier models, including both open-weights and proprietary models, to contextualize its capabilities and identify areas of strength and improvement.

\paratitle{Benchmarks}

To comprehensively assess {Ring-2.6-1T}, we conduct evaluations across a wide range of benchmarks, primarily covering 5 domains: reasoning, OpenClaw, agentic coding, agentic search, and function calling.

\begin{itemize}
    \item \textbf{Reasoning:} AIME 2026\footnote{\url{https://artofproblemsolving.com/wiki/index.php/AIME_Problems_and_Solutions}} (Avg@64), LiveCodeBench-v6~\citep{jain2025livecodebench} (2408-2505, Avg@4), GPQA-Diamond~\citep{rein2023gpqa} (Avg@16), ARC-AGI-2~\citep{chollet2025arcagi2} (Pass@2), HMMT-Feb26\footnote{\url{https://www.hmmt.org/}} (Avg@64), IMO-AnswerBench~\citep{luong-etal-2025-towards} (Avg@8).
    \item \textbf{OpenClaw:} PinchBench\footnote{\url{https://pinchbench.com/}} (Avg@3), ClawEval\footnote{ClawEval is the evaluation suite for OpenClaw agents.} (0424, Pass\textasciicircum 3).
    \item \textbf{Agentic Coding\footnote{For SWE benchmarks, we report the performance by using Claude Code as the scaffolding}:} SWE-bench Verified~\citep{jimenez2024swebench} (Resolved), SWE-bench Pro~\citep{deng2025swebenchpro} (Resolved).
    \item \textbf{Agentic Search:} GAIA-2 Search~\citep{mialon2023gaia} (Pass@1, 3 runs).
    \item \textbf{Function Calling:} $\tau^2$-bench~\citep{barres2025tau2bench} (Average, Retail, Airline, Telecom).
\end{itemize}

We evaluate {Ring-2.6-1T} under two inference configurations: {Ring-2.6-1T (xhigh)}, which uses an extended thinking budget for maximum reasoning depth, and {Ring-2.6-1T (high)}, optimized for efficiency with reduced reasoning overhead. We compare against leading models including Kimi-K2.6-Thinking, DeepSeek-V4-Pro\footnote{\url{https://huggingface.co/deepseek-ai/DeepSeek-V4-Pro}}, ChatGPT-5.4~\citep{singh2025gpt5}, Gemini-3.1-Pro\footnote{\url{https://deepmind.google/technologies/gemini/}}, GLM-5.1-Thinking~\citep{GLM51LongHorizonTasks}, Claude-Opus-4.7-Thinking\footnote{\url{https://www.anthropic.com/claude}}, and Claude-Opus-4.6-Thinking. All evaluations use controlled experimental conditions with standardized configurations.

\paratitle{Results}

Table~\ref{tab:main_results} provides a comprehensive comparison of {Ring-2.6-1T} against leading frontier models. The following sections provide a detailed analysis of its performance across different aspects.

\begin{table}[htb]
  \centering
  \caption{Performance comparison across multiple benchmarks. \textbf{Bold} indicates first place overall. \underline{Underline} indicates second place overall. $^*$ denotes results from our evaluation.}
  \label{tab:main_results}
  \resizebox{\textwidth}{!}{%
  \begin{tabular}{l|cc|ccc|cccc}
  \toprule
  \multirow{2}{*}{\textbf{Benchmark}} & \multicolumn{2}{c|}{\textbf{{Ring-2.6-1T}}} & \multicolumn{3}{c|}{Open-Weights Models} & \multicolumn{4}{c}{Proprietary Models} \\
  \cmidrule(lr){2-3} \cmidrule(lr){4-6} \cmidrule(lr){7-10}
   & xhigh & high & \textbf{{Kimi-K2.6}} & \textbf{{GLM-5.1}} & \textbf{{DS-V4-Pro}} & \textbf{\splitcell{{Claude}\\{Opus-4.6}}} & \textbf{\splitcell{{Claude}\\{Opus-4.7}}} & \textbf{\splitcell{{OpenAI}\\{GPT-5.4}}} & \textbf{\splitcell{{Gemini}\\{3.1-Pro}}} \\
  \midrule
  \multicolumn{10}{c}{\textit{Reasoning}} \\
  \midrule
  AIME 2026 $_{\text{(Avg@64)}}$ & 95.78 & 87.86 & 96.40 & 95.30 & 95.83 & 96.67 & 96.41$^*$ & \textbf{99.17} & \underline{98.33} \\
    HMMT-Feb26 $_{\text{(Avg@64)}}$ & 93.47 & 67.80 & 92.70 & -- & \underline{95.20} & 83.38$^*$ & 94.08$^*$ & \textbf{95.64}$^*$ & -- \\
  IMO-AnswerBench $_{\text{(Avg@8)}}$ & 86.12 & 66.44 & 86.00 & -- & \textbf{89.80} & 80.28$^*$ & \underline{87.31}$^*$ & 77.53$^*$ & -- \\
  LCB-v6 $_{\text{(Avg@4)}}$ & 86.95 & 76.71 & \underline{89.80} & 82.49$^*$ & -- & 85.68$^*$ & 85.46$^*$ & 81.39$^*$ & \textbf{93.28}$^*$ \\
  GPQA-Diamond $_{\text{(Avg@16)}}$ & 85.89 & 76.10 & 91.10 & 86.20 & 90.10 & 91.30& \underline{94.20} & 92.80 &
  \textbf{94.30} \\
  ARC-AGI-2 $_{\text{(Pass@2)}}$ & 66.18 & - & 29.38$^*$ & 21.74$^*$ & 62.43$^*$ & 68.80$^\dagger$ & \underline{75.80} & 74.00 &
  \textbf{77.10} \\
  \midrule
  \multicolumn{10}{c}{\textit{OpenClaw}} \\
  \midrule
  PinchBench $_{\text{(Avg@3)}}$ & -- & \textbf{87.60} & -- & 70.30$^*$ & -- & 74.95 & 75.83 & 79.95 & \underline{80.00} \\
  ClawEval $_{(\text{0424, } \text{Pass}\textasciicircum3)}$ & -- & \underline{63.82} & 62.30 & 62.30 & 59.80 & \textbf{70.40} & -- & 60.30 & 57.80 \\
  \midrule
  \multicolumn{10}{c}{\textit{Agentic Coding}} \\
  \midrule
  SWE-bench-Verified $_{\text{(Resolved)}}$ & -- & 74.00 & 80.20 & - & 80.60 & -- & \textbf{87.60} & \underline{80.60} & 80.60 \\
  SWE-bench-Pro $_{\text{(Resolved)}}$ & -- & 53.76 & -- & 58.40 & -- & -- & \textbf{64.30} & \underline{59.10} & 54.20 \\
  \midrule
  \multicolumn{10}{c}{\textit{Agentic Search}} \\
  \midrule
  GAIA-2 Search $_{\text{(Pass@1, 3 runs)}}$ & 77.90 & 75.40 & 76.04 & 75.63$^*$ & 78.33$^*$ & 73.75$^*$ & 67.92$^*$ & \textbf{82.71}$^*$ & \underline{80.83}$^*$ \\
  \midrule
  \multicolumn{10}{c}{\textit{Function Calling}} \\
  \midrule
  $\tau^2$-Average $_{\text{(Acc)}}$ & 83.44 & 84.26 & 84.28$^*$ & \textbf{88.74}$^*$ & -- & \underline{87.73}$^*$ & 85.93$^*$ & 83.63$^*$ & 84.42$^*$ \\
  $\tau^2$-Retail $_{\text{(Acc)}}$ & 77.85 & 78.07 & 72.15$^*$ & \textbf{85.09}$^*$ & -- & 81.80$^*$ & 81.80$^*$ & 74.34$^*$ & \underline{83.77}$^*$ \\
  $\tau^2$-Airline $_{\text{(Acc)}}$ & 77.50 & 78.00 & 82.00$^*$ & 82.00$^*$ & -- & \underline{82.50}$^*$ & \textbf{88.50}$^*$ & 80.50$^*$ & 76.50$^*$ \\
  $\tau^2$-Telecom $_{\text{(Acc)}}$ & 94.96 & 96.71 & 98.68$^*$ & \underline{99.12}$^*$ & -- & \textbf{99.30}$^*$ & 87.50$^*$ & 96.05$^*$ & \textbf{99.30} $^*$\\
  \bottomrule
  \end{tabular}%
  }
  \vspace{2pt}
  \end{table}

\paragraph*{Reasoning.}
{Ring-2.6-1T} demonstrates strong reasoning capabilities across challenging benchmarks. On AIME 2026, {Ring-2.6-1T (xhigh)} achieves 95.78\%, placing it competitively among the top frontier models. On LiveCodeBench-v6 (2408--2505), it scores 86.95\%, ranking second among open-weights models behind Kimi-K2.6-Thinking (89.80\%) and outperforming GLM-5.1-Thinking (82.49\%) by over 4 percentage points. On ARC-AGI-2, {Ring-2.6-1T (xhigh)} attains 66.18\%, ranking as the top open-weights model and demonstrating strong abstract reasoning and pattern recognition capabilities. These results highlight the model's robust reasoning performance, driven by our scalable reinforcement learning training recipe.

\paragraph*{OpenClaw.}
{Ring-2.6-1T} excels in tasks evaluated through the OpenClaw benchmarks. On PinchBench, {Ring-2.6-1T (high)} achieves the highest score of 87.60\% among all evaluated models, surpassing Gemini-3.1-Pro (80.00\%) and ChatGPT-5.4 (79.95\%) by a substantial margin of over 7 percentage points. On ClawEval (0424, pass$^3$), {Ring-2.6-1T (high)} achieves 63.82\%, ranking first among open-weights models and surpassing Kimi-K2.6-Thinking (62.30\%) and GLM-5.1-Thinking (62.30\%). 

\paragraph*{Agentic Coding.}
On agentic coding benchmarks, {Ring-2.6-1T} shows promising capabilities. On SWE-bench Verified, {Ring-2.6-1T (high)} achieves 74.00\%, narrowing the gap with leading open-weights models such as Kimi-K2.6-Thinking (80.20\%) and DeepSeek-V4-Pro (80.60\%). On SWE-bench Pro, it scores 53.76\%, competitive with GLM-5.1-Thinking (58.40\%). These results reflect a solid foundation for complex, real-world software engineering tasks that require multi-file reasoning and iterative debugging.

\paragraph*{Agentic Search.}
On GAIA-2 Search, {Ring-2.6-1T (xhigh)} achieves 77.90\%, demonstrating strong multi-hop reasoning and information retrieval capabilities in web search scenarios. This performance places it competitively alongside DeepSeek-V4-Pro (78.33\%) and behind the leading ChatGPT-5.4 (82.71\%) and Gemini-3.1-Pro (80.83\%). Notably, the model outperforms several strong baselines including GLM-5.1-Thinking (75.63\%) and Claude-Opus-4.6-Thinking (73.75\%), indicating effective integration of search-augmented reasoning.

\paragraph*{Function Calling.}
{Ring-2.6-1T} demonstrates robust function calling abilities across the $\tau^2$-bench suite~\citep{barres2025tau2bench}. The {high} variant achieves a $\tau^2$-Average of 84.26\%, competitive with Gemini-3.1-Pro (84.42\%) and Kimi-K2.6-Thinking (84.28\%). On $\tau^2$-Telecom, {Ring-2.6-1T (high)} scores 96.71\%, showcasing strong performance in domain-specific tool use. These results confirm the model's reliability in structured API interactions and multi-turn function calling scenarios, which are essential for real-world agentic applications.

%% file: sections/4-infrastructure.tex
\section{Infrastructure}\label{sec:infra}
The system stack is a first-order determinant of both capability and cost. In brief, the main challenge is not merely to make each component work in isolation, but to preserve throughput, numerical stability, and scheduling efficiency under trillion-parameter scale and highly variable workloads. This section therefore describes the infrastructure co-design that supports the 2.6 family across three levels: training efficiency for long-context pre-training and post-training, RL efficiency for large-scale asynchronous agentic optimization, and inference efficiency for serving.

\subsection{Long-context Training}
Long-context training in the 2.6 family is not a straightforward extension of the Ling-2.0 stack. After introducing hybrid linear attention and pushing training to much longer sequence lengths, we found that both the distributed training strategy and the memory-management policy had to be reconsidered to preserve end-to-end efficiency at scale. Our infrastructure response therefore focuses on three coupled problems: scalable context parallelism for Lightning Attention, kernel efficiency under highly fragmented variable-length workloads, and throughput-stability balancing for long-context MoE training.

\subsubsection{Context Parallel For Lightning Attention}  
Linear Attention reduces complexity from $O(N^2)$ to $O(N)$ via an RNN-style recurrence, making it a key primitive for long-context training. However, its sequential dependency along the sequence dimension precludes a direct port of conventional Context Parallel (CP) schemes designed for Softmax Attention: Ring Attention relies on online-softmax correction, which is inapplicable to recurrent state accumulation, while DeepSpeed Ulysses requires \texttt{cp\_size} to divide \texttt{head\_num}, limiting scalability under extreme context lengths.

\begin{figure}[!hbt]
  \centering
  \includegraphics[width=0.6\linewidth]{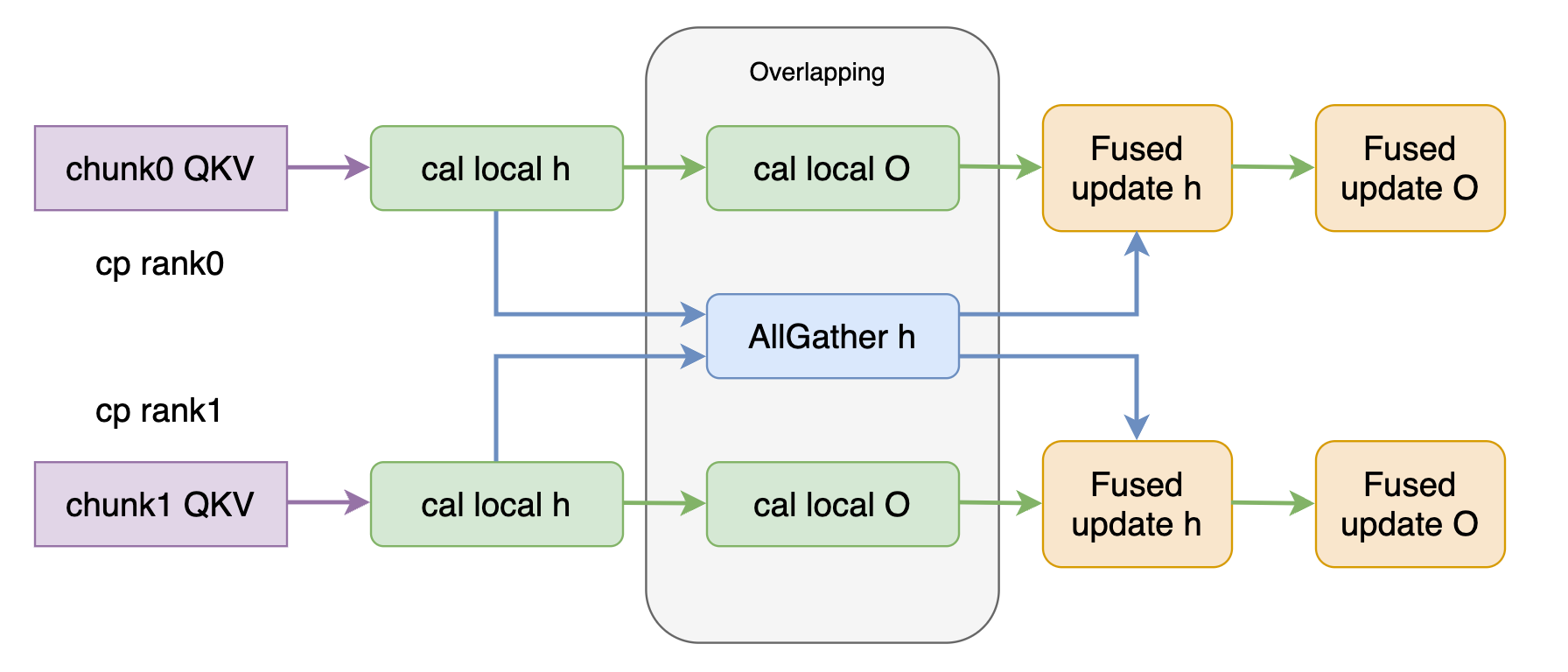}
  \caption{Lightning Attention CP Optimization.}
  \label{fig:lightning_cp}
\end{figure}

\textbf{AllGather-based CP.} We introduce an AllGather CP design for Linear Attention, built on a \emph{local-recurrence-then-global-correction} principle, as shown in Figure~\ref{fig:lightning_cp}:
\begin{itemize}
    \item \textbf{Intra-GPU stage:} Each rank independently computes its local hidden state $h^{(k)}$ and local output $O^{(k)}$ on its sequence shard, fully reusing the highly optimized FLA kernels without algorithmic modification.
    \item \textbf{Inter-GPU stage:} A single AllGather over the $[B,H,D,D]$ local states is issued; each rank then folds the preceding states along the recurrence and corrects its local output. Since FLA emits $h^{(k)}$ and $O^{(k)}$ from two separate kernels, the AllGather is overlapped with the local $O$ computation, effectively hiding the inter-rank communication cost.
\end{itemize}
This design is free of head-divisibility constraints and scales linearly with \texttt{cp\_size}, providing strictly better scalability than All2All-style alternatives for ultra-long-context regimes.

\subsubsection{Kernel Fusion for Varlen Sequences.} 
In long-context training with a large \texttt{cp\_size}, varlen inputs frequently contain a large number of short sub-sequences. Under such workloads, a naive implementation incurs frequent kernel launches that translate into substantial CPU-side overhead and markedly degraded GPU utilization, with the effect particularly pronounced in the backward pass. We mitigate this by analytically unrolling the state-correction recurrence into a vectorized matrix form and consolidating the per-sub-sequence output correction into a single Triton-fused kernel, jointly delivering an approximately 68\% end-to-end speedup at 256K context length.

\subsubsection{Throughput and Stability Balancing} 
Long-context training of MoE models creates severe imbalances in computational load and memory use across GPUs, affecting performance and stability. Key issues include router load-balancing fluctuations and uneven workloads from  block-diagonal causal masking, which worsen with context length.

Our throughput-focused strategy co-designs parallelism methods (expert, pipeline, and context parallelism) with selective activation recomputation to maximize memory use and minimize overhead. However, operating near GPU memory limits risks OOM failures during sudden router imbalance.
To manage this, we use different memory policies per training phase:
\begin{itemize}
\item \textbf{Pre-training:} Aggressive GPU memory use for higher FLOPs utilization.
\item \textbf{Post-training on long context:} Conservative GPU memory use to absorb volatility, prioritizing stability.
\end{itemize}

We also found a critical scaling issue: some existing MoE kernel implementations assume token counts fit in 32-bit integers. Expert token counts may exceed this limit in long-context training, causing overflow or crashes. We address this by auditing and modifying critical pathways to use 64-bit integers (int64) for indexing and counting, ensuring robustness at scale.

In summary, we balance high throughput and stability through phase-specific memory policies and foundational fixes, enabling reliable distributed MoE training on ultra-long context.

\subsection{Reinforcement Learning Infrastructure: ASystem}
Ling-2.6 and Ring-2.6 builds RL on ASystem, the RL framework introduced in our previous Ring-2.0~\citep{ring1t2025}. Through pluggable training, inference, and reward backends, ASystem enables independent extension and debugging of each component. By separating control flow from data flow, it effectively alleviates the single-point data-flow bottleneck found in traditional SingleController architectures. In Ling-2.6 and Ring-2.6, ASystem targets three infrastructure requirements: efficient long-sequence rollout scheduling, stable low-precision training-inference alignment, and native collection of multi-turn agentic RL trajectories.

\begin{figure}[htb]
  \centering
  \includegraphics[width=0.8\linewidth]{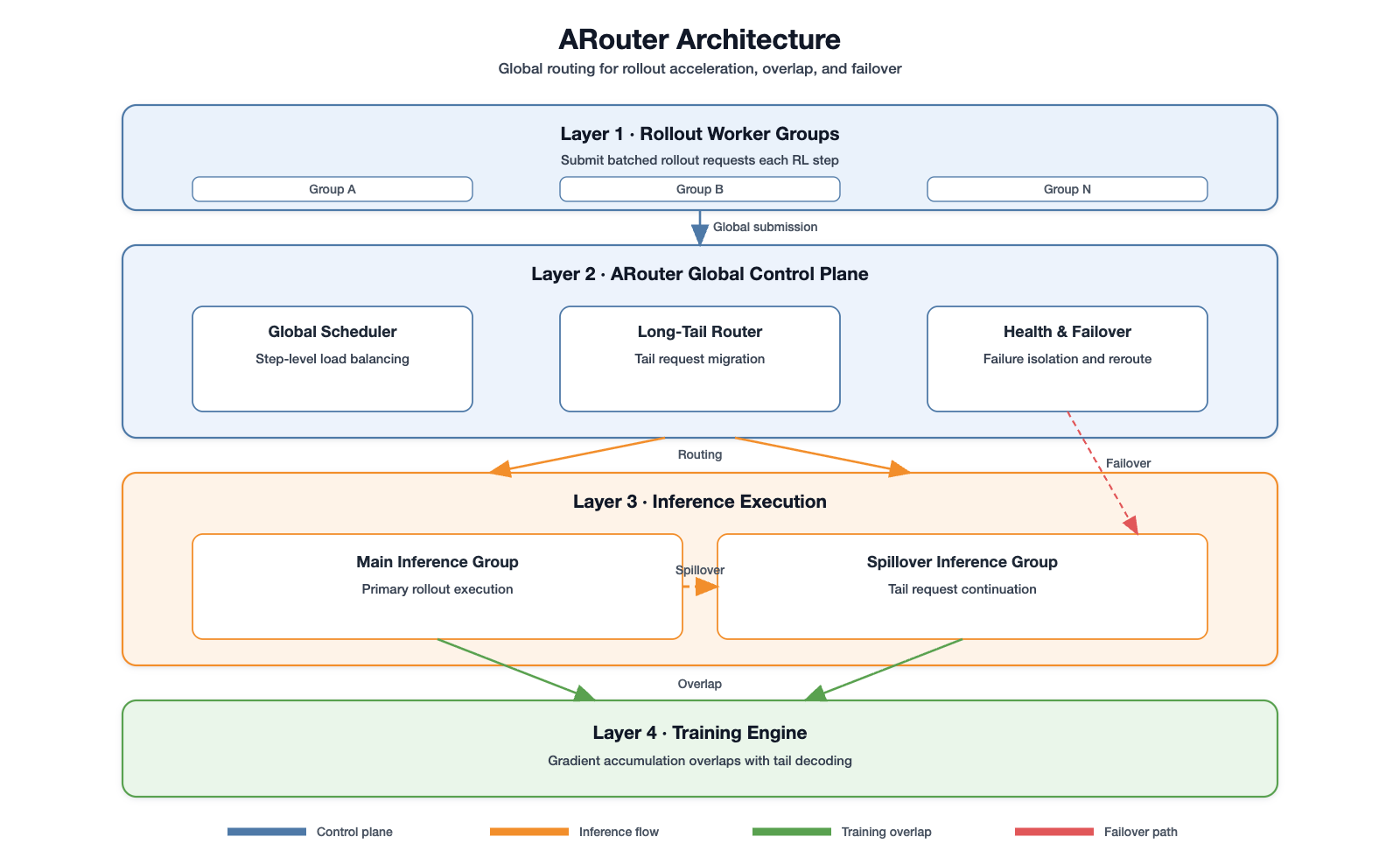}
  \caption{ARouter Architecture.}
  \label{fig:arouter_arch}
\end{figure}

\textbf{Global Rollout Scheduling with ARouter.} ARouter is the global request router for RL rollout, as shown in Figure~\ref{fig:arouter_arch}. Rather than minimizing individual request latency, it minimizes step completion time, since a small number of long decoding requests can stall an entire rollout batch and leave many GPUs idle.

To reduce tail latency, ARouter tracks inference-instance load and generation progress, then migrates late-stage tail requests from congested instances to idle ones. It also supports spillover-based training-inference overlap: residual tail requests are offloaded to a dedicated inference group, while the main inference group releases compute and starts training-side gradient accumulation. Tail results are merged before the model update, allowing training computation to be hidden inside rollout and improving end-to-end performance by more than 80\% in long-sequence scenarios.

ARouter further provides inference-instance failover. Unhealthy instances are removed from scheduling, affected requests are rerouted to healthy nodes, and request-level streaming checkpoints preserve incremental generation states. This avoids full-step reruns and reduces recovery cost for long-running rollout steps.

\textbf{Integrated FP8 Training and Inference.} ASystem integrates FP8 training and inference to improve throughput while preserving RL stability. Since Ling-2.6 uses FP8 training from pretraining through supervised fine-tuning, FP32 optimizer master weights retain higher-precision information than dequantized BF16 model weights. Loading these master weights during RL initialization improves early reward growth for both BF16 and FP8 continuation training.

To reduce training-inference mismatch, ASystem computes the LM Head in FP32 in both the Megatron training engine and the SGLang inference engine, yielding about a two-point reward improvement. For the remaining layers, ASystem uses module-aware FP8 quantization instead of applying Blockwise FP8 Linear uniformly. Attention Linear and Shared Experts Linear stay in BF16, while Routed Experts Linear uses Blockwise FP8. This preserves numerically sensitive attention computation while quantizing the MoE component that accounts for more than 90\% of parameters. In the Max 1T 128K RL setting, this design controls log-probability drift, matches the BF16 baseline in long-running evaluations, and improves end-to-end throughput by 30\%.

\textbf{Agentic RL Support.} ASystem supports multi-turn agentic RL by decoupling agent execution from distributed inference and training. Agents call a standard OpenAI Chat Completions API over HTTP, while a proxy layer routes requests to SGLang, records interactions, and builds trajectory trees for training. Multiple independent sessions can run for each prompt, with session-level fault isolation so a failed agent run does not block the batch.

To support training-signal organization in multi-turn agent interactions, ASystem supports individual mode, which trains each turn with loss only on the current response, while concat mode trains the full trajectory with terminal reward. For complex agents, it detects retries and exploration branches, extracts the main chain, and can optionally train on branches or share rewards, focusing learning on critical decisions.

\input{sections/reasoning/async}

\begin{figure}[htb]
  \centering
  \includegraphics[width=0.8\linewidth]{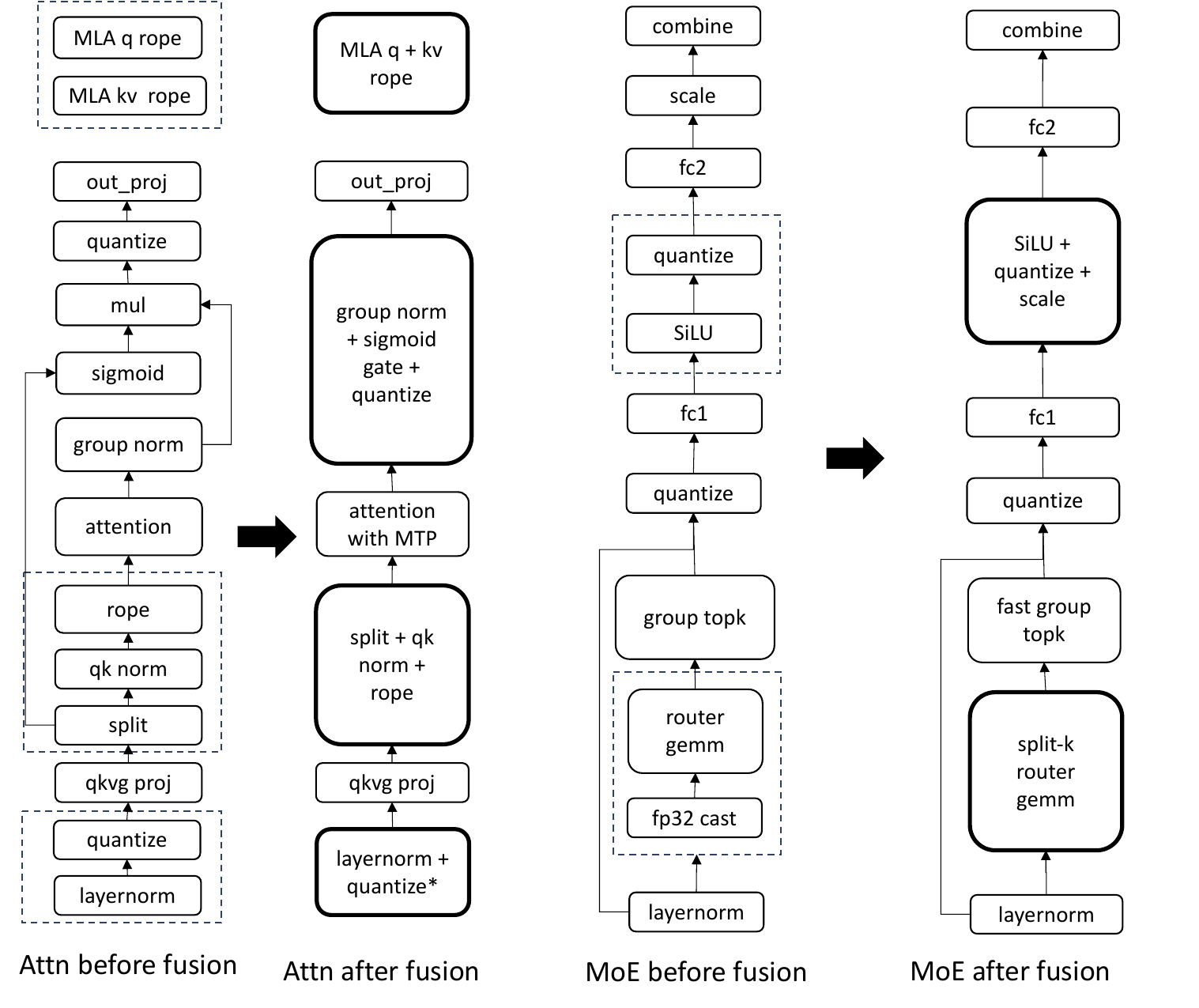}
  \caption{Inference optimization with linghe.}
  \label{fig:linghe_infer}
\end{figure}

\begin{table}[htb]
\centering
\begin{tabular}{lccc}
\toprule
 & Baseline (tokens/s) & MTP (tokens/s) & MTP+linghe (tokens/s) \\
\midrule
BF16 (BS=1)  & 186  & 257 (+38\%)  & 297 (+60\%)  \\
BF16 (BS=16) & 1075 & 1114 (+4\%)  & 1233 (+15\%) \\
FP8 (BS=1)   & 155  & 236 (+51\%)  & 341 (+119\%) \\
FP8 (BS=16)  & 1078 & 1173 (+9\%)  & 1664 (+54\%) \\
\bottomrule
\end{tabular}
\caption{Throughput comparison under different precision and batch size settings. The input length is 16,384, with an 8,192-token shared prefix.}
\label{tab:throughput_comparison}
\end{table}

\subsection{Operator Fusion}
During the pretraining of Ling-2.6, we significantly improved training efficiency through extensive operator fusion. On the inference side, we further adapted these fused kernels for real deployment scenarios, aligning their fusion granularity and numerical behavior as closely as possible with the training stage. This design not only improves inference efficiency, but also reduces training-inference divergence during RL rollout. These inference kernels has been open-sourced through \textbf{linghe}\footnote{https://github.com/inclusionAI/linghe}. To support different precision settings, we performed systematic optimization across the inference stage.

\textbf{BF16 inference:} We implemented fusion for key operators such as QK Norm + RoPE and Group RMSNorm + Sigmoid Gate. We also adopted a BF16 Input + FP32 Output computation path for both MoE Router GEMM and LM Head GEMM, while further optimizing the implementations of MLA RoPE and Top-K.

\textbf{FP8 inference:} We further fused RMS Norm, SwiGLU, with quantization, and introduced Split-K Blockwise FP8 GEMM for small-batch scenarios to unlock additional throughput gains.

Taken together, these optimizations form a system-level co-design across kernel fusion, prefix caching mechanisms, and multi-token generation. The result is not just higher overall system throughput, but also higher per-user TPS, shorter wait times, and a more stable, fluid interactive experience in real-world usage.

%% file: sections/reasoning/async.tex
\subsubsection{Asynchronous RL}
Ring-2.6 further extends ASystem with asynchronous RL execution to handle long-tailed reasoning rollouts and environment-bound agentic workloads. Reasoning tasks such as math, coding, logic, and STEM can produce chains of thought whose tail generations are much longer than the average. Agentic tasks such as coding, search, and tool use introduce additional latency from code sandboxes, terminal sessions, retrieval services, and tool/MCP endpoints. A synchronous lock-step RL pipeline would leave the trainer waiting for the slowest rollout. ASystem therefore focuses on continuously feeding the trainer while bounding the policy staleness introduced by asynchronous execution.

\textbf{Partial-Rollout Pipeline.} Rather than fully decoupling rollout and training, Ring-2.6 RL runs in a \emph{partial-rollout} regime in which each optimization step is gated by a global token budget on rollout generation, in the spirit of C3PO++~\citep{ring1t2025}. A \textbf{rollout controller} continuously dispatches prompts to the inference engine through an \emph{overlap batch dispatcher}, which submits new work while previous trajectories are still being collected so that the inference pool is never drained at a batch boundary. As the trajectories arrive, the controller tracks the cumulative number of generated tokens against the per-iteration budget $\Phi$. Completed trajectories are routed through ASandbox for verifier or tool-driven reward computation; trajectories that are still in flight when the budget is reached are \emph{paused}, persisted into a cross-version rollout buffer together with their KV-cache fingerprint, and resumed by the next policy version in a subsequent iteration. The trainer then performs one optimization step on the harvested batch and triggers an update of weights on the inference engines. By limiting each step based on token counts instead of the slowest-running trajectory, this approach avoids the long-tail overhead that a fully synchronous rollout would incur, while still keeping the boundary between rollout and training deterministic and easy to checkpoint.

Because trajectories may now be assembled from segments generated by different policy versions, we manage the resulting version skew through a dedicated \textbf{staleness manager}. Every rollout segment is tagged with the inference engine's policy version at the time of generation; the manager admits new rollouts into the pool only up to a configurable bound of $\texttt{max\_staleness} \times \texttt{consumer\_batch\_size}$, and discards or retires segments whose version gap to the trainer exceeds this bound. The result is a bounded-staleness regime in which the RL algorithm can treat the rollouts as approximately on-policy, while the system extracts the throughput of partial-rollout execution. Together with ASystem's sub-second weight broadcasts, the effective staleness during the reinforcement learning of Ring-2.6 is limited to only a few optimization steps, and we observe no loss in stability or final reward compared to the synchronous baseline.

\textbf{Asynchronous Agentic Rollouts.} The partial-rollout pipeline is what allows us to run agentic RL in a tractable way at our scale. For coding tasks, the trajectories operate within a containerized repository, invoking tools for building, testing, and shell access; for search tasks, they perform retrieval, browsing, and reasoning via calls to external services; for tool-use tasks, they interact with a multi-turn customer-service simulator whose state is tracked outside of the inference engine. In each case, the rollout is a sequence of interactions (model generate, environment step, model generate, $\dots$) whose per-step latency is highly variable and dominated by the environment rather than by the GPU.

Three properties of the stack make these workloads efficient. First, the inference engine and the environments are \emph{scaled independently}: thousands of ASandbox replicas handle varied environments, and the rollout controller models each tool invocation as an awaitable, ensuring that a GPU is never held up waiting on a pending HTTP/MCP request. Second, the overlap batch dispatcher \emph{continuously refills} the inference engine as soon as any in-flight trajectory yields a partial result, while the token-budget cutoff bounds the cost of pathologically long trajectories by placing them in the rollout buffer for the next policy version; together, these mechanisms ensure that the long tail of a code patch or a browsing session no longer leaves the GPU idle. Third, the unified verifier interface covering diverse reasoning tasks, the coding harness, the tool-use simulator, and the browsers, all presented through a single HTTP/MCP contract under ASandbox that allows the trainer to combine reasoning and agentic datasets within a single run, without needing task-specific scheduling.

%% file: sections/6-conclusion.tex
\section{Conclusion, Limitations, and Future Directions}
In this report, we presented Ling-2.6 and Ring-2.6, a model family for practical agentic intelligence at trillion-parameter scale. 
This model family illustrates that progress at this scale should come less from model size alone than from tighter co-design across architecture, post-training, systems, and agent training environments. Within this framework, this report advances three goals: efficient long-context processing, higher capability per output token, and more reliable environment-grounded agentic behavior. Ling-2.6 is developed as an instant model line optimized for fast response, strong token efficiency, and broad utility under deployment constraints, whereas Ring-2.6 is developed as a deeper-thinking model line for complex reasoning and long-horizon agentic tasks. Taken together, the 2.6 family should be understood not simply as a new set of models, but as a concrete path toward efficient, open, and increasingly practical-oriented agentic systems.

Despite these advances, several important bottlenecks remain unresolved. Ling-2.6-flash gains substantial throughput and token economy, but its tighter deliberation budget necessarily constrains reasoning depth, instruction compositionality, and tool-use reliability in high-complexity settings. The current token-efficiency objective is also incomplete. It compresses procedural reasoning effectively, yet it does not always cleanly separate low-value repetition from necessary factual elaboration in knowledge-intensive outputs. Furthermore, long-horizon agentic robustness is still weaker than short-horizon competence: the models can make strong local decisions, but reliability degrades over extended workflows, shifting tool states, and heterogeneous execution environments.  Finally, alignment and evaluation remain less stable than desired: the models can drift across languages under highly constrained prompts, and existing public benchmarks still under-measure persistence, recovery behavior, cost-aware planning, and deployment-time robustness.

The next stage of Ling and Ring should continue this co-design perspective while pushing the family toward higher capability ceilings. In brief, further gains will depend on deeper efficiency co-design across architecture and systems, including model scaling, advanced architectural designs, low-precision training and inference, KV-cache management, and next-generation optimization recipes such as Muon~\citep{liu2025muon}.
An equally important transition is from text-only systems to native multimodal agents. Practical agents must operate over visual interfaces, documents, code artifacts, and mixed-modality environments, so future Ling and Ring models should extend the current architecture and agentic RL stack to native multimodality rather than rely on loosely coupled external perception modules. Evaluation must also evolve in the same direction. Taken together, 
these directions define a compact research agenda for the our next stage agentic systems: larger and more capable models, tighter efficiency co-design across the stack, and more deployment-grounded multimodal agents.

%% file: sections/author.tex
\clearpage
\section{Contributors}
\label{sec:contri}

\DTLnewdb{names}
\DTLnewrow{names} \DTLnewdbentry{names}{name}{Jun Zhou$^{\dagger}$} 
\DTLnewrow{names} \DTLnewdbentry{names}{name}{Zhiqiang Zhang$^{\dagger}$} 
\DTLnewrow{names} \DTLnewdbentry{names}{name}{Zujie Wen} 
\DTLnewrow{names} \DTLnewdbentry{names}{name}{Haoyu Xu} 
\DTLnewrow{names} \DTLnewdbentry{names}{name}{Binbin Hu} 
\DTLnewrow{names} \DTLnewdbentry{names}{name}{Fangzheng Zhao} 
\DTLnewrow{names} \DTLnewdbentry{names}{name}{Bing Li} 
\DTLnewrow{names} \DTLnewdbentry{names}{name}{Shuo Zhang} 
\DTLnewrow{names} \DTLnewdbentry{names}{name}{Ben Liu} 
\DTLnewrow{names} \DTLnewdbentry{names}{name}{Feifan Wu} 
\DTLnewrow{names} \DTLnewdbentry{names}{name}{Hao Qian} 
\DTLnewrow{names} \DTLnewdbentry{names}{name}{Kunlong Chen} 
\DTLnewrow{names} \DTLnewdbentry{names}{name}{Wenjing Fang} 
\DTLnewrow{names} \DTLnewdbentry{names}{name}{Shaowei Wei} 
\DTLnewrow{names} \DTLnewdbentry{names}{name}{Changxin Tian} 
\DTLnewrow{names} \DTLnewdbentry{names}{name}{Qian Zhao} 
\DTLnewrow{names} \DTLnewdbentry{names}{name}{Qingyuan Yang} 
\DTLnewrow{names} \DTLnewdbentry{names}{name}{Lanyin Mei} 
\DTLnewrow{names} \DTLnewdbentry{names}{name}{Ziqi Liu} 
\DTLnewrow{names} \DTLnewdbentry{names}{name}{Jiapeng Wang} 
\DTLnewrow{names} \DTLnewdbentry{names}{name}{Liang Jiang} 
\DTLnewrow{names} \DTLnewdbentry{names}{name}{Shuaicheng Li} 
\DTLnewrow{names} \DTLnewdbentry{names}{name}{Zhengyu Huang} 
\DTLnewrow{names} \DTLnewdbentry{names}{name}{Pan Tang} 
\DTLnewrow{names} \DTLnewdbentry{names}{name}{Xiaopei Wan} 
\DTLnewrow{names} \DTLnewdbentry{names}{name}{Hongrui Liu} 
\DTLnewrow{names} \DTLnewdbentry{names}{name}{Dingyuan Zhu} 
\DTLnewrow{names} \DTLnewdbentry{names}{name}{Meiqi Zhu} 
\DTLnewrow{names} \DTLnewdbentry{names}{name}{Shujie Li} 
\DTLnewrow{names} \DTLnewdbentry{names}{name}{Tong Zhao} 
\DTLnewrow{names} \DTLnewdbentry{names}{name}{Weiguang Han} 
\DTLnewrow{names} \DTLnewdbentry{names}{name}{Mingyuan Fan} 
\DTLnewrow{names} \DTLnewdbentry{names}{name}{Daixin Wang} 
\DTLnewrow{names} \DTLnewdbentry{names}{name}{Xinyu Kong} 
\DTLnewrow{names} \DTLnewdbentry{names}{name}{Qianggang Cao} 
\DTLnewrow{names} \DTLnewdbentry{names}{name}{Tianshu Wang} 
\DTLnewrow{names} \DTLnewdbentry{names}{name}{Yiqi Wang} 
\DTLnewrow{names} \DTLnewdbentry{names}{name}{Zhenduo Zhang} 
\DTLnewrow{names} \DTLnewdbentry{names}{name}{Jia Liu} 
\DTLnewrow{names} \DTLnewdbentry{names}{name}{Cunyin Peng} 
\DTLnewrow{names} \DTLnewdbentry{names}{name}{Peijie Jiang} 
\DTLnewrow{names} \DTLnewdbentry{names}{name}{Zihang Zeng} 
\DTLnewrow{names} \DTLnewdbentry{names}{name}{Yuxin Tian} 
\DTLnewrow{names} \DTLnewdbentry{names}{name}{Hailin Zhao} 
\DTLnewrow{names} \DTLnewdbentry{names}{name}{Yingjie Song} 
\DTLnewrow{names} \DTLnewdbentry{names}{name}{Ke Shi} 
\DTLnewrow{names} \DTLnewdbentry{names}{name}{Zihao Wang} 
\DTLnewrow{names} \DTLnewdbentry{names}{name}{Kuan Xu} 
\DTLnewrow{names} \DTLnewdbentry{names}{name}{Jia Guo} 
\DTLnewrow{names} \DTLnewdbentry{names}{name}{Guodong Yang} 
\DTLnewrow{names} \DTLnewdbentry{names}{name}{Nuo Xu} 
\DTLnewrow{names} \DTLnewdbentry{names}{name}{Ye Chen} 
\DTLnewrow{names} \DTLnewdbentry{names}{name}{Zhidong Fan} 
\DTLnewrow{names} \DTLnewdbentry{names}{name}{Peilong Zhao} 
\DTLnewrow{names} \DTLnewdbentry{names}{name}{Yuxiao He} 
\DTLnewrow{names} \DTLnewdbentry{names}{name}{Mengjie Gao} 
\DTLnewrow{names} \DTLnewdbentry{names}{name}{Huaidong Xiong} 
\DTLnewrow{names} \DTLnewdbentry{names}{name}{Xuan Sun} 
\DTLnewrow{names} \DTLnewdbentry{names}{name}{Kun Tang} 
\DTLnewrow{names} \DTLnewdbentry{names}{name}{Jin Yang} 
\DTLnewrow{names} \DTLnewdbentry{names}{name}{Jiannan Shi} 
\DTLnewrow{names} \DTLnewdbentry{names}{name}{Dalong Zhang} 
\DTLnewrow{names} \DTLnewdbentry{names}{name}{Chengyao Wen} 
\DTLnewrow{names} \DTLnewdbentry{names}{name}{Qiang Cheng} 
\DTLnewrow{names} \DTLnewdbentry{names}{name}{Bin Jing} 
\DTLnewrow{names} \DTLnewdbentry{names}{name}{Hanxiao Zhang} 
\DTLnewrow{names} \DTLnewdbentry{names}{name}{Fanzhuang Meng} 
\DTLnewrow{names} \DTLnewdbentry{names}{name}{Zhengke Gui} 
\DTLnewrow{names} \DTLnewdbentry{names}{name}{Zhizhen Liu} 
\DTLnewrow{names} \DTLnewdbentry{names}{name}{Lei Liang} 
\DTLnewrow{names} \DTLnewdbentry{names}{name}{Deng Zhao} 
\DTLnewrow{names} \DTLnewdbentry{names}{name}{Jiaming Liu} 
\DTLnewrow{names} \DTLnewdbentry{names}{name}{Cai Chen} 
\DTLnewrow{names} \DTLnewdbentry{names}{name}{Xiaodong Yan} 
\DTLnewrow{names} \DTLnewdbentry{names}{name}{Bin Hu} 
\DTLnewrow{names} \DTLnewdbentry{names}{name}{Liangcheng Fu} 
\DTLnewrow{names} \DTLnewdbentry{names}{name}{Qi Zuo} 
\DTLnewrow{names} \DTLnewdbentry{names}{name}{Dingnan Jin} 
\DTLnewrow{names} \DTLnewdbentry{names}{name}{Shaomian Zheng} 
\DTLnewrow{names} \DTLnewdbentry{names}{name}{Chen Qian} 
\DTLnewrow{names} \DTLnewdbentry{names}{name}{Jianping Wei} 
\DTLnewrow{names} \DTLnewdbentry{names}{name}{Juanhui Tu} 
\DTLnewrow{names} \DTLnewdbentry{names}{name}{Zilong Wang} 
\DTLnewrow{names} \DTLnewdbentry{names}{name}{Lihui Zhang} 
\DTLnewrow{names} \DTLnewdbentry{names}{name}{Jie Wang} 
\DTLnewrow{names} \DTLnewdbentry{names}{name}{Lu Yu} 
\DTLnewrow{names} \DTLnewdbentry{names}{name}{Junpeng Fang} 
\DTLnewrow{names} \DTLnewdbentry{names}{name}{Zhigang Huangfu} 
\DTLnewrow{names} \DTLnewdbentry{names}{name}{Jiaolong Yang} 
\DTLnewrow{names} \DTLnewdbentry{names}{name}{Yue Gao} 
\DTLnewrow{names} \DTLnewdbentry{names}{name}{Yeyang Chen} 
\DTLnewrow{names} \DTLnewdbentry{names}{name}{Tiange Xu} 
\DTLnewrow{names} \DTLnewdbentry{names}{name}{Haoxiong Liu} 
\DTLnewrow{names} \DTLnewdbentry{names}{name}{Qing Cui} 
\DTLnewrow{names} \DTLnewdbentry{names}{name}{Junnan Sipan} 
\DTLnewrow{names} \DTLnewdbentry{names}{name}{Xinyao Tang} 
\DTLnewrow{names} \DTLnewdbentry{names}{name}{Yuxiao Lu} 
\DTLnewrow{names} \DTLnewdbentry{names}{name}{Sikang Bian} 
\DTLnewrow{names} \DTLnewdbentry{names}{name}{Junjie Ou} 
\DTLnewrow{names} \DTLnewdbentry{names}{name}{Zhaoxin Huan} 
\DTLnewrow{names} \DTLnewdbentry{names}{name}{Xiao Shi} 
\DTLnewrow{names} \DTLnewdbentry{names}{name}{Jinquan Sun} 
\DTLnewrow{names} \DTLnewdbentry{names}{name}{Qiyin Huang} 
\DTLnewrow{names} \DTLnewdbentry{names}{name}{Cong Zhang} 
\DTLnewrow{names} \DTLnewdbentry{names}{name}{Zheping Qu} 
\DTLnewrow{names} \DTLnewdbentry{names}{name}{Xiaqing Sun} 
\DTLnewrow{names} \DTLnewdbentry{names}{name}{Xiong Xu} 
\DTLnewrow{names} \DTLnewdbentry{names}{name}{Hanzi Wang} 
\DTLnewrow{names} \DTLnewdbentry{names}{name}{Hao Liu} 
\DTLnewrow{names} \DTLnewdbentry{names}{name}{Mingming Yin} 
\DTLnewrow{names} \DTLnewdbentry{names}{name}{Yuqi Zhang} 
\DTLnewrow{names} \DTLnewdbentry{names}{name}{Wanli Gu} 
\DTLnewrow{names} \DTLnewdbentry{names}{name}{Zitao Xuan} 
\DTLnewrow{names} \DTLnewdbentry{names}{name}{Ang Li} 
\DTLnewrow{names} \DTLnewdbentry{names}{name}{Pingping Liu} 
\DTLnewrow{names} \DTLnewdbentry{names}{name}{Chunwei Wu} 
\DTLnewrow{names} \DTLnewdbentry{names}{name}{Kaiqin Hu} 
\DTLnewrow{names} \DTLnewdbentry{names}{name}{Chilin Fu} 
\DTLnewrow{names} \DTLnewdbentry{names}{name}{Zuoli Tang} 
\DTLnewrow{names} \DTLnewdbentry{names}{name}{Xiaolu Zhang} 
\DTLnewrow{names} \DTLnewdbentry{names}{name}{Weichang Wu} 
\DTLnewrow{names} \DTLnewdbentry{names}{name}{Longfei Zheng} 
\DTLnewrow{names} \DTLnewdbentry{names}{name}{Zhihao Wang} 
\DTLnewrow{names} \DTLnewdbentry{names}{name}{Quan Wan} 
\DTLnewrow{names} \DTLnewdbentry{names}{name}{Yingxue Li} 
\DTLnewrow{names} \DTLnewdbentry{names}{name}{Yuxuan Li} 
\DTLnewrow{names} \DTLnewdbentry{names}{name}{Shengnan Zhang} 
\DTLnewrow{names} \DTLnewdbentry{names}{name}{Zhankai Xu} 
\DTLnewrow{names} \DTLnewdbentry{names}{name}{Hongzhi Ruan} 
\DTLnewrow{names} \DTLnewdbentry{names}{name}{Linfeng Shi} 
\DTLnewrow{names} \DTLnewdbentry{names}{name}{Longfei Li} 
\DTLnewrow{names} \DTLnewdbentry{names}{name}{Yalin Zhang} 
\DTLnewrow{names} \DTLnewdbentry{names}{name}{Xinxing Yang} 
\DTLnewrow{names} \DTLnewdbentry{names}{name}{Meng Li} 
\DTLnewrow{names} \DTLnewdbentry{names}{name}{Caizhi Tang} 
\DTLnewrow{names} \DTLnewdbentry{names}{name}{Yankun Ren} 
\DTLnewrow{names} \DTLnewdbentry{names}{name}{Yue Zhang} 
\DTLnewrow{names} \DTLnewdbentry{names}{name}{Bin Han} 
\DTLnewrow{names} \DTLnewdbentry{names}{name}{Yixuan Sun} 
\DTLnewrow{names} \DTLnewdbentry{names}{name}{Donghao Zhang} 
\DTLnewrow{names} \DTLnewdbentry{names}{name}{Yao Zhao} 
\DTLnewrow{names} \DTLnewdbentry{names}{name}{Chen Liang} 
\DTLnewrow{names} \DTLnewdbentry{names}{name}{Wenbo Shen} 
\DTLnewrow{names} \DTLnewdbentry{names}{name}{Zibin Lin} 
\DTLnewrow{names} \DTLnewdbentry{names}{name}{Mingyang Zhang} 
\DTLnewrow{names} \DTLnewdbentry{names}{name}{Yibo Cao} 
\DTLnewrow{names} \DTLnewdbentry{names}{name}{Jie Gao} 
\DTLnewrow{names} \DTLnewdbentry{names}{name}{Zixuan Cheng} 
\DTLnewrow{names} \DTLnewdbentry{names}{name}{Fan Yuan} 
\DTLnewrow{names} \DTLnewdbentry{names}{name}{Man Li} 
\DTLnewrow{names} \DTLnewdbentry{names}{name}{Jiameng Wang} 
\DTLnewrow{names} \DTLnewdbentry{names}{name}{Qizheng Zhou} 
\DTLnewrow{names} \DTLnewdbentry{names}{name}{Hao Wu} 
\DTLnewrow{names} \DTLnewdbentry{names}{name}{Hongxun Li} 
\DTLnewrow{names} \DTLnewdbentry{names}{name}{Wei Lu} 
\DTLnewrow{names} \DTLnewdbentry{names}{name}{Wenzhi Tang} 
\DTLnewrow{names} \DTLnewdbentry{names}{name}{Yongfei Xu} 
\DTLnewrow{names} \DTLnewdbentry{names}{name}{Kaihong Zhang} 
\DTLnewrow{names} \DTLnewdbentry{names}{name}{Yue Yu} 
\DTLnewrow{names} \DTLnewdbentry{names}{name}{Jinjin Li} 
\DTLnewrow{names} \DTLnewdbentry{names}{name}{Tingwei Huang} 
\DTLnewrow{names} \DTLnewdbentry{names}{name}{Huihuang Zheng} 
\DTLnewrow{names} \DTLnewdbentry{names}{name}{Haitao Zhang} 
\DTLnewrow{names} \DTLnewdbentry{names}{name}{Lei Chen} 
\DTLnewrow{names} \DTLnewdbentry{names}{name}{Jun Liu} 
\DTLnewrow{names} \DTLnewdbentry{names}{name}{Jinjing Huang} 
\DTLnewrow{names} \DTLnewdbentry{names}{name}{Xinyu Liu} 
\DTLnewrow{names} \DTLnewdbentry{names}{name}{Huikang Tang} 
\DTLnewrow{names} \DTLnewdbentry{names}{name}{Fengbin Fang} 
\DTLnewrow{names} \DTLnewdbentry{names}{name}{Weiquan Li} 
\DTLnewrow{names} \DTLnewdbentry{names}{name}{Ying Li} 
\DTLnewrow{names} \DTLnewdbentry{names}{name}{Chao Huang} 
\DTLnewrow{names} \DTLnewdbentry{names}{name}{Peng Lin} 
\DTLnewrow{names} \DTLnewdbentry{names}{name}{Yuzhuo Fu} 
\DTLnewrow{names} \DTLnewdbentry{names}{name}{Zhuyan Zhou} 
\DTLnewrow{names} \DTLnewdbentry{names}{name}{Wang Hong} 
\DTLnewrow{names} \DTLnewdbentry{names}{name}{Gangshan Wang} 
\DTLnewrow{names} \DTLnewdbentry{names}{name}{Xin Zhao} 
\DTLnewrow{names} \DTLnewdbentry{names}{name}{Jinyao Chen} 
\DTLnewrow{names} \DTLnewdbentry{names}{name}{Xudong Wang} 
\DTLnewrow{names} \DTLnewdbentry{names}{name}{Yizhu Xiao} 
\DTLnewrow{names} \DTLnewdbentry{names}{name}{Jiewei Wu} 
\DTLnewrow{names} \DTLnewdbentry{names}{name}{Chao Zhang} 
\DTLnewrow{names} \DTLnewdbentry{names}{name}{Li Tang} 
\DTLnewrow{names} \DTLnewdbentry{names}{name}{Ting Guo} 
\DTLnewrow{names} \DTLnewdbentry{names}{name}{Tinghao Wang} 
\DTLnewrow{names} \DTLnewdbentry{names}{name}{Chengfu Tang} 
\DTLnewrow{names} \DTLnewdbentry{names}{name}{Feng Xu} 
\DTLnewrow{names} \DTLnewdbentry{names}{name}{Liyuan Liu} 
\DTLnewrow{names} \DTLnewdbentry{names}{name}{Zhenjun Ma} 
\DTLnewrow{names} \DTLnewdbentry{names}{name}{Yijie Chen} 
\DTLnewrow{names} \DTLnewdbentry{names}{name}{Runyuan Zhao} 
\DTLnewrow{names} \DTLnewdbentry{names}{name}{Xudong Han} 
\DTLnewrow{names} \DTLnewdbentry{names}{name}{Hong Liu} 
\DTLnewrow{names} \DTLnewdbentry{names}{name}{Tongkai Yang} 
\DTLnewrow{names} \DTLnewdbentry{names}{name}{Hao Dai} 
\DTLnewrow{names} \DTLnewdbentry{names}{name}{Xujie Shen} 
\DTLnewrow{names} \DTLnewdbentry{names}{name}{Haitao Wang} 
\DTLnewrow{names} \DTLnewdbentry{names}{name}{Jun Mei} 
\DTLnewrow{names} \DTLnewdbentry{names}{name}{Zhenxuan Pan} 
\DTLnewrow{names} \DTLnewdbentry{names}{name}{Xingyu Lu} 
\DTLnewrow{names} \DTLnewdbentry{names}{name}{Qitao Shi} 
\DTLnewrow{names} \DTLnewdbentry{names}{name}{Qiaoben Bao} 
\DTLnewrow{names} \DTLnewdbentry{names}{name}{Xiang Shu} 
\DTLnewrow{names} \DTLnewdbentry{names}{name}{Lintao Ma} 
\DTLnewrow{names} \DTLnewdbentry{names}{name}{Zhibo Zhu} 
\DTLnewrow{names} \DTLnewdbentry{names}{name}{Yu Liu} 
\DTLnewrow{names} \DTLnewdbentry{names}{name}{Jia Li} 
\DTLnewrow{names} \DTLnewdbentry{names}{name}{Yifan Zuo} 
\DTLnewrow{names} \DTLnewdbentry{names}{name}{Tianchu Yao} 
\DTLnewrow{names} \DTLnewdbentry{names}{name}{Yuchen Fang} 
\DTLnewrow{names} \DTLnewdbentry{names}{name}{Lei Xu} 
\DTLnewrow{names} \DTLnewdbentry{names}{name}{Heng Zhang} 
\DTLnewrow{names} \DTLnewdbentry{names}{name}{Lu Liu} 
\DTLnewrow{names} \DTLnewdbentry{names}{name}{Hongliang Zhang} 
\DTLnewrow{names} \DTLnewdbentry{names}{name}{Mengshu Sun} 
\DTLnewrow{names} \DTLnewdbentry{names}{name}{Yangyang Hou} 
\DTLnewrow{names} \DTLnewdbentry{names}{name}{Jun Xu} 

\DTLsort{name}{names}

\large{Contributors are listed \textbf{alphabetically by the first name}.} 

\large{
\begin{multicols}{3}
\raggedcolumns
\DTLforeach*{names}{\thename=name}{\thename\\}
\end{multicols}}

$^{\dagger}$ denotes corresponding authors.

\clearpage

%% file: sections/8-appendix.tex
\appendix
\section{Agentic Coding Environments}
\label{sec:agent_env}

\begin{figure}[!hbt]
  \centering
  \includegraphics[width=0.6\linewidth]{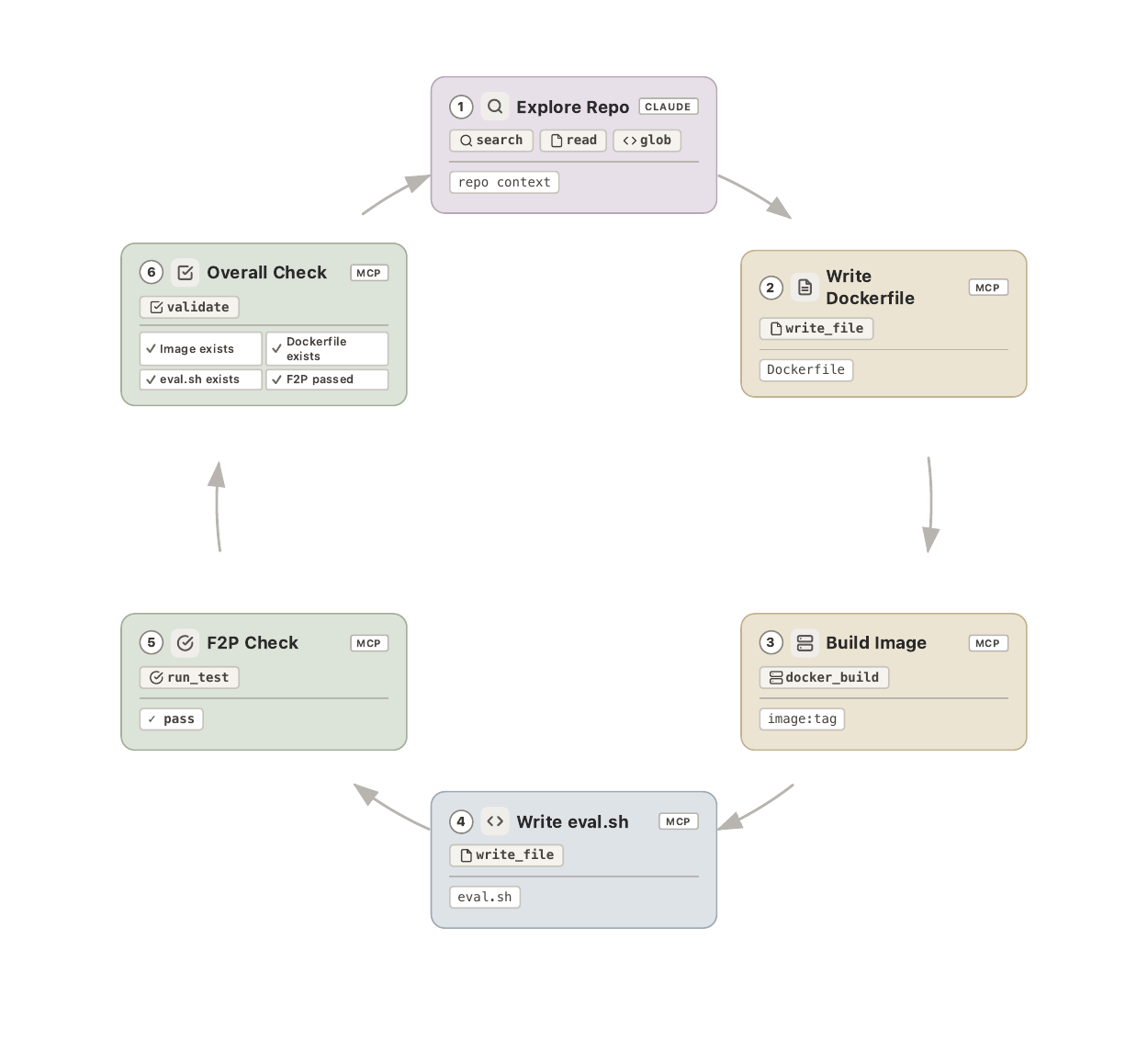}
  \caption{Construction workflow of coding environments.}
  \label{fig:swe-flowchart}
\end{figure}

Agentic coding tasks typically run in an isolated environment where the agent interacts with files and system utilities to solve a given problem. Thus reproducible execution environments are essential for both training and evaluation, as they provide the infrastructure required for faithful rollout and verification. We adopt a Claude Code + MCP approach to automatically generate Docker images for each instance. The key idea is to leverage Claude Code to explore a repository's codebase and documentation, gather sufficient context, and produce a working Dockerfile. To mitigate hallucination and ensure reliability, the LLM is constrained to interact with the environment solely through a designated set of MCP tools. For instance, image builds may only be triggered via the `build\_image` tool, which copies a clean repository snapshot into the image through the Dockerfile `COPY` instruction and guarantees that the in-image repository remains unmodified. Similarly, the `verify\_task\_completion` tool checks that all required artifacts, including the Dockerfile, evaluation script, and built image, are present and that Fail2Pass verification passes. Figure~\ref{fig:swe-flowchart} illustrates the construction workflow. By combining the flexibility of LLM-driven exploration with rule-based tool validation, we rapidly and reliably produced approximately 220,000 Docker images.

\section{Multi-Token Prediction with Continued Training}
\label{sec:mtp_continue}
Beyond SFT data refinement and RL optimization, Ling-2.6 further improves inference efficiency during continued training through multi-token prediction (MTP). MTP can improve the performance of the base model and can also serve as a draft model for speculative decoding, thereby substantially accelerating inference. However, a standard MTP layer~\citep{gloeckle2024better, deepseekai2024deepseekv3technicalreport} is trained to predict only the next token. When it is used to predict multiple tokens during inference, a discrepancy between training and inference arises~\citep{leviathan2023fast}, which reduces the accepted length. To alleviate this issue, multiple MTP layers must be introduced during training. We therefore incorporate two additional MTP layers during the post-training stage and continue training the MTP layers.

As shown in Table \ref{tab:mtp}, under the same number of speculative steps, namely four, on our private evaluation dataset, the MTP model after continued training achieves a moderate improvement in accepted length compared with the standard MTP model. We also observe that using only the first MTP layer to predict all subsequent tokens yields a notable increase in accepted length. This suggests that the newly introduced MTP layers may remain insufficiently trained in the continued-training setting. To further test this hypothesis, we share parameters across different MTP layers~\citep{zeng2026glm} and detach the gradients from all MTP layers except the first one, thereby preventing them from propagating to the base model. The results show that this strategy further improves the accepted length during inference while reducing memory overhead.

\begin{table}[htbp]
\centering
\large
\caption{Comparison of accept lengths of different MTP architecture.}
\label{tab:mtp}
\renewcommand{\arraystretch}{1.2}
\begin{tabular}{l|cccc}
\toprule
 & MTP-1 & MTP-3 & MTP-3-1 & MTP-3-share \\
\midrule
Acc Length & 2.71 & 3.23 & 3.29 & \textbf{3.31} \\
\bottomrule
\end{tabular}
\end{table}